\documentclass[journal,transmag]{IEEEtran}
\usepackage{url}
\usepackage{amssymb,amsmath}
\usepackage{subfigure}
\usepackage{multirow}
\usepackage{times,epsfig}
\usepackage{dblfloatfix}
\usepackage[vlined,ruled,linesnumbered]{algorithm2e}
\usepackage{epsfig,tabularx,amssymb,amsmath,subfigure,multirow,
algorithmic,graphicx}
\usepackage{color,graphicx}
\usepackage{subfigure,multirow}
\usepackage{times}
\usepackage{amsmath}
\usepackage{epsfig}
\usepackage{ctable}
\usepackage{threeparttable}
\usepackage{enumerate}
\usepackage{amsmath}
\usepackage{amsfonts}
\usepackage{color}
\usepackage{multirow}
\usepackage{graphicx}
\usepackage{balance}
\usepackage{paralist}
\usepackage{CJKutf8}
\usepackage{url}
\usepackage{amssymb}
\usepackage{multirow}

\begin{document}
\title{Transfer Learning-Based Outdoor Position Recovery \\with Telco Data}

\author{Yige Zhang, Aaron Yi Ding, J\"{o}rg Ott, Mingxuan Yuan, Jia Zeng, Kun Zhang and Weixiong Rao
\IEEEcompsocitemizethanks{
\IEEEcompsocthanksitem Yige Zhang and Weixiong Rao are with Tongji University, Shanghai, China. \protect
E-mail: \{1610832, wxrao\}@tongji.edu.cn
\IEEEcompsocthanksitem Aaron Yi Ding is with the Department of Engineering Systems and Services at TU Delft, Netherlands.
\protect
E-mail: Aaron.Ding@tudelft.nl
\IEEEcompsocthanksitem J\"{o}rg Ott is with Technical University of Munich in the Faculty
of Informatics.
\protect
E-mail: ott@in.tum.de
\IEEEcompsocthanksitem Mingxuan Yuan and Jia Zeng are with Huawei Noahs Ark Lab, Hong Kong. \protect
E-mail: \{mingxuan.yuan, jia.zeng\}@huawei.com
\IEEEcompsocthanksitem Kun Zhang is with the CMU philosophy department as an assistant professor and an affiliate faculty member in the machine learning department.
\protect
E-mail: kunz1@andrew.cmu.edu
}
\thanks{}}

\maketitle
\begin{abstract}
Telecommunication (Telco) outdoor position recovery aims to localize outdoor mobile devices by leveraging measurement report (MR) data. Unfortunately, Telco position recovery requires sufficient amount of MR samples across different areas and suffers from high data collection cost. For an area with scarce MR samples, it is hard to achieve good accuracy. In this paper, by leveraging the recently developed transfer learning techniques, we design a novel Telco position recovery framework, called \textsf{TLoc}, to transfer good models in the carefully selected source domains (those fine-grained small subareas) to a target one which originally suffers from poor localization accuracy. Specifically, \textsf{TLoc} introduces three dedicated components: 1) a new coordinate space to divide an area of interest into smaller domains, 2) a similarity measurement to select best source domains, and 3) an adaptation of an existing transfer learning approach. To the best of our knowledge, \textsf{TLoc} is the first framework that demonstrates the efficacy of applying transfer learning in the Telco outdoor position recovery. To exemplify, on the 2G GSM and 4G LTE MR datasets in Shanghai, \textsf{TLoc} outperforms a non-transfer approach by 27.58\% and 26.12\% less median errors, and further leads to 47.77\% and 49.22\% less median errors than a recent fingerprinting approach NBL.
\end{abstract}

\section{Introduction}\label{sec:introduction}
Recent years we have witnessed the ever-growing size and complexity of telecommunication (Telco) networks to process 1000-fold growth in the amount of traffic and 100-fold increase in the number of users \cite{MargoliesBBDJUV17}. Telco operators have to manage heterogeneous networks (including 2G-4G and upcoming 5G networks), composed of macro cells, small cells, and distributed antenna systems. The growing demands and heterogeneous networks require an automated approach to network control and management, instead of error-prone manual network management and parameter configuration. To enable the automated network control and management, the \emph{outdoor locations} of mobile devices are important for Telco operators to \emph{1}) pinpoint location hotspots for capacity planning, \emph{2}) identify gaps in radio frequency spatial coverage, and \emph{3}) locate users in emergency situations (E911) \cite{MargoliesBBDJUV17}. Moreover, the locations of mobile devices are widely used to understand mobility patterns and optimize many third-party applications such as urban planning and traffic forecasting \cite{BeckerCHLUVV11}.

Outdoor locations of mobile devices can be recovered from Telco Measurement record (MR) data \cite{CostaZ18}. MR samples are generated when mobile devices make phone calls and access data services. MR samples contain connection states (e.g., signal strength) between mobile devices and connected base stations. After the locations of mobile devices are recovered, we tag the MR samples by the associated geo-locations, generating the so-called geo-tagged MR samples.

In literature, various position recovery algorithms via Telco MR samples have been developed. Google MyLocation \cite{googlemylocation} approximates outdoor locations by the positions of cellular towers connected with mobile devices. This method suffers from median errors of hundreds and even thousands of meters. More recently, data-driven approaches have attracted intensive research interests in both academia and Telco industry \cite{AlyY13,IbrahimY12,Koshima354477,ThiagarajanRBMG10,YuanDZLNHWDY14,ZhuLYZZGDRZ16}. These approaches leverage geo-tagged MR samples to build the mapping from MR samples to associated locations, and the mapping is then used to localize the mobile devices in non-geo-tagged samples. For example, the fingerprinting approach \cite{IbrahimY12} builds a histogram of MR signal strength (i.e., fingerprint database) for each divided grid cell in the areas of interest, and the Random Forest (RaF)-based approach \cite{ZhuLYZZGDRZ16} maintains the mapping function between MR features (i.e., MR signal strength) and position labels. When enough amount of training geo-tagged MR samples are used, the data-driven algorithms achieve the median error of $20 \thicksim 80$ meters \cite{DBLP:conf/mdm/HuangRZLYZY17,ZhuLYZZGDRZ16}.%

A key concern of the data-driven methods mentioned above is requiring sufficient geo-tagged MR samples to build the accurate mapping from MR samples to associated locations. Nevertheless, collecting sufficient geo-tagged MR samples across the distributed areas of an urban city incurs rather high cost. It is not rare that an area of interest suffers from insufficient geo-tagged MR samples. If we have scarce geo-tagged MR samples for such an area, the position recovery precision in that area could be very low. For example in a recent work NBL \cite{MargoliesBBDJUV17}, though over 100 TB GPS-tagged Telco signal data in an American city are collected by 4 million users from Jan 2016 to July 2016, the median localization errors in rural areas are still as high as around 750 meters due to insufficient samples.

In this paper, targeting Telco operators, we design a transfer learning-based Telco position recovery approach, called \textsf{TLoc}, to accurately localize mobile devices in those areas with scarce data samples. The general idea of \textsf{TLoc} is as follows. First, we divide an entire area into fine-grained small subareas, namely \emph{domains}. For each domain, we then maintain the mapping from MR samples within this domain to their associated positions. Next for the target domains suffering from low precision, we transfer good mappings from appropriately selected source domains to target ones via transfer learning. In this way, we greatly improve the localization accuracy in target domains. Though transfer learning has been used for indoor WiFi-based localization  \cite{DBLP:conf/aaai/PanSYK08,DBLP:conf/aaai/ZhengPYP08,DBLP:conf/aaai/ZhengXYS08}, we believe that indoor WiFi and outdoor Telco localization differs  significantly. Thus, those indoor localization approaches are not expected to perform well in our case (they will be evaluated in our experiment) due to the following challenges. \emph{Firstly}, given the outdoor Telco localization, designing a proper position coordinate space is the prerequisite to enable knowledge transfer across two domains. This, unfortunately, cannot be achieved by using outdoor GPS longitude and latitude coordinates: the different GPS position (i.e., position label) for every area makes it impossible to share position labels across distributed domains, and hence hard to perform knowledge transfer. \emph{Secondly}, given a large number of domains, it is challenging to select the best source domains for a target one. In contrast, due to the small area and rather limited domains in an indoor environment, it is straightforward for indoor localization to select source domains. Thus, trivial effort on source domain selection is employed for indoor WiFi-based localization  \cite{DBLP:conf/aaai/PanSYK08,DBLP:conf/aaai/ZhengPYP08,DBLP:conf/aaai/ZhengXYS08}.

To tackle the challenges above, \textsf{TLoc} builds the following components. Firstly, unlike absolute GPS coordinates, we use a \emph{relative coordinate space} for position recovery. Under this coordinate space, the mobile devices even in two distributed domains can still share the same relative positions, facilitating the transfer across two domains. Secondly, based on the relative position space, we design an effective distance metric to measure the similarity between domains. The metric incorporates the distribution of the signal strength of MR samples, relative position information, and non-serving base station deployment information. Finally, by adapting an existing structured transfer learning (STL) method \cite{DBLP:journals/pami/SegevHMCE17}, we build a Random Forest (RaF)-based position recovery model for each domain and then perform model transfer from appropriately chosen source domains to target ones. As a summary, this paper makes the following contributions.
\begin{itemize}
\item To the best of our knowledge, \textsf{TLoc} is the first method to plausibly leverage transfer learning for Telco outdoor localization. Unlike the fingerprinting-based and machine learning-based approaches \cite{DBLP:conf/mdm/HuangRZLYZY17,Koshima354477,ZhuLYZZGDRZ16}, \textsf{TLoc} mitigates high efforts to collect a large quantity of training samples across an entire area. Moreover, our evaluation empirically verifies that the idea of \textsf{TLoc} can generally benefit other approaches (e.g., fingerprinting-based approaches) to achieve better precision by re-using MR samples from source domains to target ones. 
\item We design a novel approach to divide an entire urban area into small domains by the proposed relative coordinate space. Based on the divided domains, we define a distance metric for measuring domain similarity to select appropriate source domains effectively for a target one. By adapting a recent structured transfer learning (STL) scheme \cite{DBLP:journals/pami/SegevHMCE17} for a RaF regression model \cite{ZhuLYZZGDRZ16}, \textsf{TLoc} leads to much better position recovery precision than those non-transfer models.
\item Our extensive evaluation validates that \textsf{TLoc} greatly outperforms both state-of-the-arts and the variants of \textsf{TLoc}. For example, on two 2G GSM and 4G LTE MR datasets, \textsf{TLoc} outperforms the recent fingerprinting approach NBL \cite{MargoliesBBDJUV17} by 47.77\% and 49.22\% less median errors, respectively, and leads to 27.58\% and 26.12\% less median error when compared with the non-transfer RaF algorithm \cite{ZhuLYZZGDRZ16}, respectively.
\end{itemize}

The rest of this paper is organized as follows. Section \ref{s:problem} reviews the background and related work. Section \ref{s:design} gives the general idea of \textsf{TLoc} and the proposed relative coordinate space. After that, Section \ref{sec:source} defines the distance metric to measure domain similarity, and Section \ref{sec:transfer} adapts the STL model \cite{DBLP:journals/pami/SegevHMCE17} for \textsf{TLoc}. Section \ref{s:evaluation} evaluates \textsf{TLoc} and Section \ref{s:conclusion} finally concludes the paper. Table \ref{fig:symbols} summarizes the main acronyms and notations used in the paper.

\begin{table}[!ht]\scriptsize 
\centering
\begin{tabular}{|l|l|} \hline
\textbf{Notation/Symbol} & \textbf{Meaning}\\ \hline\hline
Telco & Telecommunication \\
MR& Measurement Report \\  
TL & Transfer Learning\\
STL&Structure Transfer Learning\\
SVR &Supported Vector Regression \\RSSI &Received Signal Strength Indicator\\
RaF &Random Forest\\\hline\hline
$D$, $s$ & Domain (divided small areas), MR Sample \\
$\mathbf{S}_T$ and $\mathbf{S}_S$ & Target and Source Data Set \\
$F_d()$, $F_i()$ &MR Features dependent (resp. independent) upon locations\\
$L()$& Recovered Location \\
$dis_{mr}^{rssi}$, $dis_{mr}^{sig}$  & Weighted Histogram Distance of RSSI (resp. SignalLevel) \\
$dis_{mr}$ &Overall MR Feature Distance \\ $dis_{pos}$ & Relative Position Distance \\
$dist(D,D')$ & Domain Distance between two domains $D$ and $D'$ \\\hline
\end{tabular}
\caption{ Mainly used short names/symbols and notations.}\label{fig:symbols}
\end{table}

\section{Background and Related Work} \label{s:problem}
In this section, firstly we give the background of Measurement Report (MR), Random Forests (RaF), and transfer learning, and secondly review the literature in terms of outdoor position recovery and selection of source domains.

\begin{table}[hbt]
\scriptsize\hspace{-3ex}
\centering
\begin{tabular}{|l|l|l|l|l|}
\hline
MRTime xxx&	IMSI xxx&	RNCID 6188&	BestCellID 26050&	NumBS 7\\\hline
RNCID$_1$ 6188&	CellID$_1$ 26050&	AsuLevel$_1$ 18&	SignalLevel$_1$ 4&	RSSI$_1$ -77\\ \hline
RNCID$_2$ 6188&	CellID$_2$ 27394&	AsuLevel$_2$ 16&	SignalLevel$_2$ 4&	RSSI$_2$ -81\\ \hline
RNCID$_3$ 6188&	CellID$_3$ 27377&	AsuLevel$_3$ 15&	SignalLevel$_3$ 4&	RSSI$_3$ -83 \\ \hline
RNCID$_4$ 6188& CellID$_4$ 27378&	AsuLevel$_4$ 15&	SignalLevel$_4$ 4&	RSSI$_4$ -83\\ \hline
RNCID$_5$ 6182&	CellID$_5$ 41139&	AsuLevel$_5$ 16&	SignalLevel$_5$ 4&  RSSI$_5$ -89\\ \hline
RNCID$_6$ 6188&	CellID$_6$ 27393&	AsuLevel$_6$ 9&	    SignalLevel$_6$ 3& RSSI$_6$ -95\\ \hline
RNCID$_7$ 6182&	CellID$_7$ 26051&	AsuLevel$_7$ 9&	   SignalLevel$_7$ 3&	RSSI$_7$ -95\\ \hline
\end{tabular}
\caption{A 2G GSM MR sample collected by an Android device.}\label{tab:mr}
\end{table}

\textbf{Measurement Report (MR) Data}: MR samples are used to record the connection states between mobile devices and nearby base stations in a Telco network. Table \ref{tab:mr} gives an example of 2G GSM MR samples collected by an Android device. It contains a unique number (IMSI: International Mobile Subscriber Identity), connection time stamp (MRTime), up to 7 nearby base stations (identified by RNCID and CellID) \cite{Rizk2019CellinDeep}, signal measurements such as AsuLevel and SignalLevel,  and a radio signal strength indicator (RSSI). SignalLevel indicates the power ratio (typically logarithm value) of the output signal of the device and the input signal. AsuLevel, i.e., Arbitrary Strength Unit (ASU), is an integer value proportional to the received signal strength measured by the mobile phone. Among the up-to 7 base stations, one of them is selected as the primary serving base station to provide communication and data transmission services for the mobile device. Previous work on Telco localization \cite{DBLP:conf/mdm/HuangRZLYZY17,ZhuLYZZGDRZ16} might ignore the use of serving base station. Unlike these works, we will carefully exploit serving base stations as the base of \textsf{TLoc}.

Besides 2G GSM MR samples, we also collect 4G LTE MR samples by frontend Android devices. They both follow the  same data format. Nevertheless, due to the limitation of Android API, frontend Android devices cannot acquire the identifiers (RNCID\_$2 \sim 7$ and CellID\_$2 \sim 7$) of non-serving base stations from 4G LTE networks. Nevertheless, the signal measurements associated with the missed base stations can be still collected.

Finally, Telco operators can collect MR samples via backend base stations except the frontend MR samples above by Android mobile phones. Nevertheless, their data formats are different \cite{DBLP:conf/mdm/HuangRZLYZY17}. Firstly, besides RSSI, the backend 4G MR samples provided by Telco operators contain RSRP and RSRQ which do not appear in the frontend 4G MR samples. Secondly, backend 2G MR samples contain RxLev, the received signal strength on ARFCN (Absolute Radio Frequency Channel Number) \cite{DBLP:conf/mdm/HuangRZLYZY17}. The previous work \cite{hoy2014forensic} shows that RxLev is exactly equal to the RSSI value, and we thus treat RxLev equally as RSSI. Now, to make sure that we have proper knowledge transfer between frontend and backend MR datasets, we perform knowledge transfer only for those MR feature items (e.g., RSSI) that appear within all datasets. For example, we transfer the knowledge from the RSSI (or RxLev) items in backend 2G MR samples to the RSSI items in frontend 2G samples. Yet, we do not transfer knowledge for such MR features as RSRP and RSRQ.

\textbf{Random Forest} (RaF) is an ensemble method for classification, regression, and other learning tasks. It constructs a multitude of {decision trees} (DTs) \cite{urldt} during the training phase and outputs either the class that is the mode of the classes (classification) or mean prediction (regression) of the individual trees. RaF avoids the overfitting of DTs to their training set. Specifically, DTs that are grown very deep tend to learn highly irregular patterns: they overfit their training sets, i.e., low bias but very high variance. RaFs are a way of averaging multiple deep DTs, trained on different parts of the same training set, with the goal of reducing the variance. This greatly boosts the performance in the final model, at the expense of a small increase in the bias and loss of interpretability.

\textbf{Transfer learning} aims at improving the learning in a new task through proper transfer of knowledge from a related task that has already been learned. Those machine learning algorithms such as RaFs are designed to address a \emph{single} task. In contrast, transfer learning attempts to leverage individual tasks by developing methods to transfer knowledge learned in one or more source tasks to a related target task. Transfer learning is frequently used due to expensive cost or impossibility to re-collect the needed training data and rebuild the models. Transfer learning approaches include \emph{Model Transfer}, \emph{Instance Transfer} (or data sample transfer), \emph{Features Transfer}, and \emph{Relational knowledge-Transfer}. We refer interested readers to the detailed survey of transfer learning \cite{PanY10}.

\textsf{TLoc} mainly utilizes a recent model transfer scheme, i.e., the structure transfer learning (STL) \cite{DBLP:journals/pami/SegevHMCE17} in decision tree (DT)-based model to transfer knowledge from multiple source domains to the target one. Specifically, DTs for similar problems (in various domains) exhibit a certain extent of structural similarity. However, the scale of the features used to construct RaF and their associated decision thresholds are likely to differ from various problems. Thus, the DTs trained on source domains are adapted to the target one by discarding all numeric threshold values in the original DTs and working top-down, and then selecting a new threshold for a node with a numeric feature using the subset of target examples that reach this node.

Recall that general transfer learning frameworks (such as instance transfer) require training examples from source domains for domain adaptation. Instead, the STL scheme can directly adapt the already trained models from source domains to target ones. This unique property is particularly useful for the scenario that cannot directly leverages source examples for domain adaption, for whatever reason, e.g., storage capacity or data privacy. Thus, \textsf{TLoc} can comfortably adapt a given source model to a target domain relying on a relatively small training set from the target. The experimental results show that multi-source transfer in STL has the better precision than the single-source transfer. 


\textbf{Outdoor position recovery}: In the literature, Telco outdoor position recovery techniques are broadly classified into two categories: \emph{1}) measurement-based methods \cite{Caffery1998}, \emph{2}) data-driven methods. Measurement-based methods frequently adopt absolute point-to-point distance estimation or angle estimation from Telco signals to calculate mobile device locations. Examples of measurement-based techniques include Angle of arrival (AOA), Time of Arrival (TOA) \cite{DBLP:journals/wpc/ChangC12}, and Received signal strength (RSS)-based single source localization \cite{DBLP:conf/icassp/VaghefiGS11}. Nevertheless, information related to AOA and TOA is highly error prone in cellular systems, and measurement-based techniques suffer from high localization errors, typically with the median error of hundreds of meters \cite{DBLP:journals/wpc/ChangC12,RayDM16}. In addition, as shown in previous work \cite{RayDM16}, 4G LTE MR samples typically have signal strength from at most two cells, namely, the serving cell and the strongest neighboring cell. Triangulation-based localization approaches thus do not perform well because they require signal strength from at least three cells.

Fingerprinting-based and machine learning-based algorithms generally belong to the data-driven methods. They both leverage collected historical data samples for outdoor position recovery. \emph{Fingerprinting methods} \cite{IbrahimY12,Koshima354477} have been reported to have better performance than measurement-based approaches. For example, in the offline survey phase, the classic work, CellSense \cite{IbrahimY12}, first divides the area of interest into smaller cells and constructs a fingerprint database, e.g., a vector histogram of RSSI on each cell. When given a query (i.e., an input RSSI feature), the online prediction phase then searches the fingerprint database to find the location that has the maximum probability given the received signal strength vector in the query. An average of the $k$ most probable fingerprint cells, weighted by the probability of each location, can be used to obtain a better estimate of locations. In addition, a better CellSense-hybrid technique consists of two phases: the rough estimation phase first uses the standard probabilistic fingerprinting technique to obtain the most probable cell in which a user may be located, and the estimation refinement phase then uses a $k$-nearest neighbor approach to estimate the closest fingerprint point, in the signal strength space, to the current user location inside the cell estimated in phase one.

AT\&T researchers recently studied the fingerprinting-based outdoor localization problems \cite{RayDM16,MargoliesBBDJUV17}. In particular, the authors in NBL \cite{MargoliesBBDJUV17} extended CellSense \cite{IbrahimY12} similarly using two stages. In an offline stage, NBL developed radio frequency coverage maps based on a large-scale crowdsourced channel measurement campaign. Then, in an online stage, a localization algorithm quickly matches the input radio frequency measurements to coverage maps. By assuming a Gaussian distribution of signal strength within each divided grid, NBL maintains the mean value and standard deviation of signal strength of each neighboring cell tower for the samples in the grid. The online stage computes the predicted location by using either Maximum Likelihood Estimation (MLE) or Weighted Average (WA). The median errors in the 4G LTE network reported by NBL are around 80 and 750 meters in urban and rural areas, respectively.

\emph{Machine learning approaches} leverage machine learning models such as Random Forests, Support Vector Regression (SVR), Gradient Boosting Decision Tree (GBDT), and artificial neural network (ANN) to build the mapping from MR features (which are extracted from MR samples and engineering parameter data of connected base stations) to device positions \cite{ZhuLYZZGDRZ16,DBLP:conf/mdm/HuangRZLYZY17}. When given an MR record without position information, machine learning models then predict the associated location. As shown in \cite{ZhuLYZZGDRZ16}, the authors proposed a context-aware coarse-to-fine regression (CCR) model (implemented by a two-layer RaFs). The CCR model takes as input 258 dimensional coarse features and 34 dimensional fine-grained contextual feature vectors. Thus, beyond strength indicators frequently used by fingerprinting approaches, those context-aware and coarse-to-fine features such as moving speed enable CCR to outperform the classic fingerprinting approaches with slightly 14\% lower median errors. In a very recent deep learning-based outdoor cellular localization system, namely DeepLoc \cite{DBLP:conf/gis/ShokryTY18}, a data augmentor is used to handle data noise issue and to provide more training samples. With help of the samples, a deep learning model is trained for better localization result.

\textbf{Source Domain Selection}: Given a number of diverse source domains, to successfully perform knowledge transfer, one may need to select a certain number of source domains that bear essential similarity to the considered target domain. Some previous works in transfer learning studied the general source domain selection problem. For example, an information theoretic framework was developed \cite{DBLP:journals/pr/AfridiRS18} to rank source convolution neural networks (CNNs) and select the top-$k$ CNNs for the target learning task by understanding the source-target relationship. A restricted boltzman machine (RBM) was also used \cite{Ammar2014An} to select source domains in the context of reinforcement learning. Some works instead did not take domain selection into account and focused on instance selection from available source domains  \cite{DBLP:journals/prl/LinAZ13}. In addition, many existing transfer learning methods suppose that source domains are provided in advance by default.

Compared to the above methods, \textsf{TLoc} gives a meaningful distance metric to determine the domain similarities for Telco position recovery task. Unlike \textsf{TLoc}, the previous work \cite{DBLP:journals/pr/AfridiRS18} focuses on the selection of pre-trained CNN models which can be intuitively treated as learning tasks, and \cite{DBLP:journals/prl/LinAZ13} selects source instances.

\section{System Design}\label{s:design}

\subsection{General Idea}\label{sec:gidea}

\begin{figure}[th]
\begin{center}					
\centerline{\includegraphics[height=3cm]{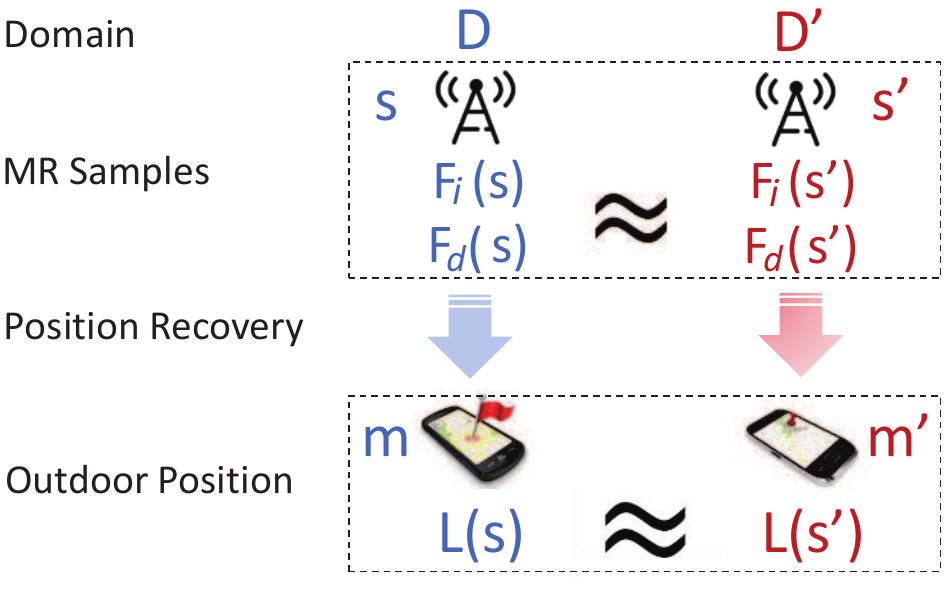}}\vspace{-2ex}
\caption{General Idea of TLoc} \label{fig:idea}\vspace{-4ex}
\end{center}
\end{figure}

We first give the general idea of \textsf{TLoc} to perform model transfer across different domains. In Figure \ref{fig:idea}, we consider that two mobile devices $m$ and $m'$ in two distributed domains $D$ and $D'$ (with $D\neq D{'}$) generate the MR samples $s$ and $s'$, respectively. Suppose that we are using a RaF regression model to recover the outdoor locations $L(s)$ and $L(s')$ for the samples $s$ and $s'$, respectively. The outdoor locations are frequently represented by GPS coordinates \cite{DBLP:conf/mdm/HuangRZLYZY17,MargoliesBBDJUV17,RayDM16,ZhuLYZZGDRZ16}.
Given the two distributed domains $D\neq D{'}$, the MR samples $s$ and $s'$ within the two domains indicate that the corresponding \emph{RNC}/\emph{CellID} and GPS positions are different, indicating $s\neq s'$ and $L(s) \neq L(s')$.

Inside MR samples, we note that there exist two types of features: 1) those ID-alike features $F_d()$ \underline{d}ependent upon located domains such as \emph{RNCID} and \emph{CellID}, and 2) those numeric features $F_i()$ \underline{i}ndependent upon located domains such as \emph{AsuLevel}, \emph{SignalLevel} and \emph{RSSI}. Due to the distributed domains $D$ and $D'$, $F_d(s)\neq F_d(s')$ holds. Nevertheless, when the two samples $s$ and $s'$ contain very similar Telco signal strength (including \emph{AsuLevel}, \emph{SignalLevel} and \emph{RSSI}), it is highly possible to have $F_i(s)\simeq F_i(s')$. The similar Telco signal strength gives us a hint: we would like to modify the representation of the features $F_d()$ and locations $L()$ to ensure that $F_d(s)\simeq F_d(s')$ and  $L(s) \simeq L(s')$ hold. When both $F_i(s)\simeq F_i(s')$ and $F_d(s)\simeq F_d(s')$ hold, we then could have the similar MR samples $s\simeq s'$ and the roughly equal positions $L(s) \simeq L(s')$. Based on this representation, we next perform knowledge transfer across two similar MR samples $s\in D \simeq s'\in D'$ as follows: if $s\simeq s'$ holds, we estimate the position $L(s) \gets L(s')$ via the position $L(s')$. In general, we extend the idea of \textsf{TLoc} from similar samples to similar domains. Gvien the two similar domains $D\simeq D'$, we infer $s\in D \simeq s'\in D'$, and then estimate $L(s) \gets L(s')$ via the available position $L(s')$.

\subsection{Relative Coordinate Space}\label{sec:relative}
To perform the knowledge transfer above, we introduce a relative coordinate space to represent $L()$ and $F_d()$, such that $F_i(s)\simeq F_i(s')$ and $F_d(s)\simeq F_d(s')$ hold for two samples $s$ and $s'$ within two distributed domains $D$ and $D'$.

\begin{figure}[th]
\begin{center}					
\centerline{\includegraphics[height=4.0cm]{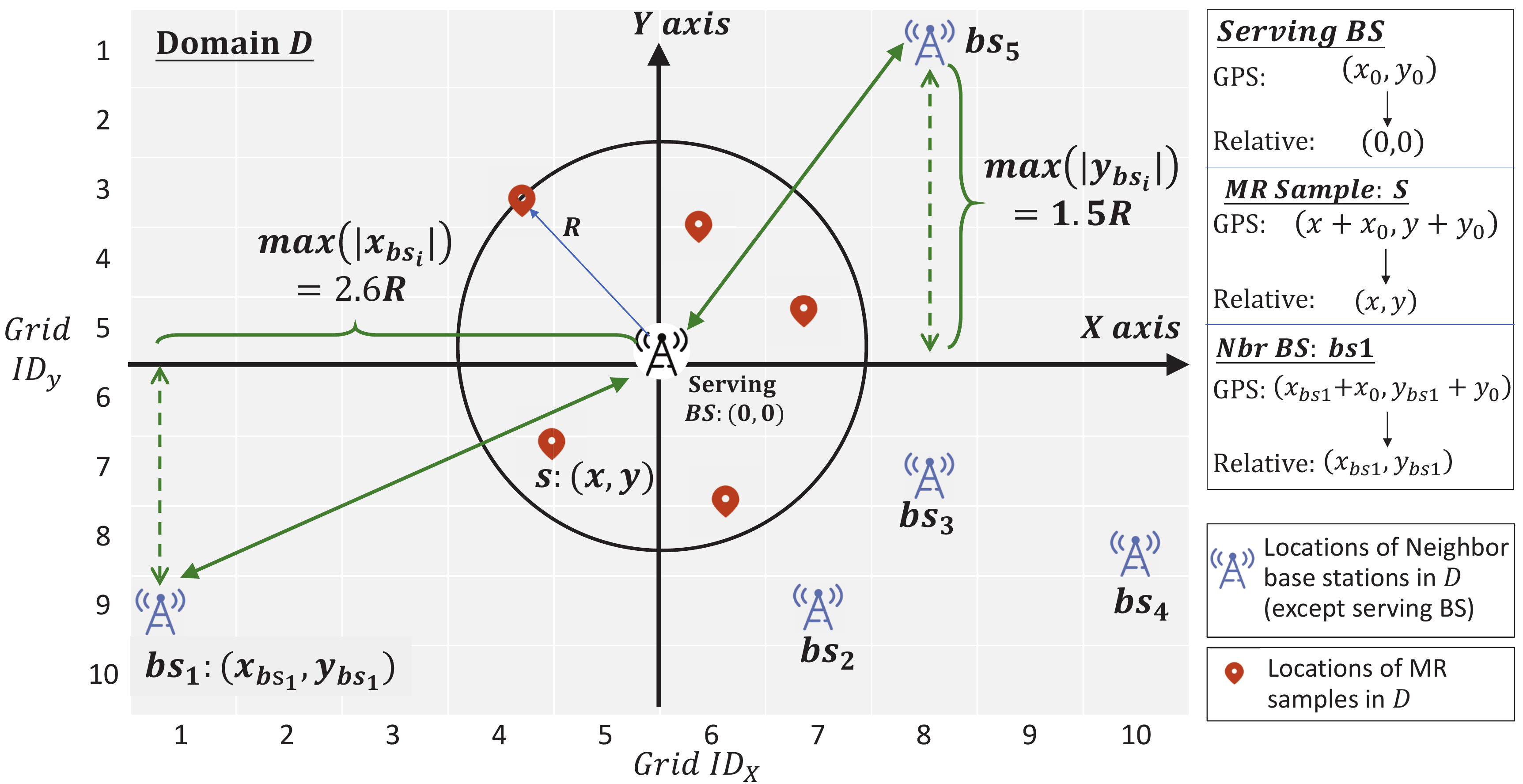}}\vspace{-2ex}
\caption{Relative Coordinate Space} \label{fig:relative_transfer}
\end{center}\vspace{-4ex}
\end{figure}

1) \textbf{Representation of $L()$}: We first represent $L()$ by transforming original GPS coordinates to relative coordinates as follows. For the MR samples  having a certain base station as their serving stations, the mobile devices generating such MR samples are highly possible to be located around the serving base station. Thus, based on serving base stations, we divide a large urban area of interest (e.g., either a university campus or an entire city) into fine-grained \emph{small subareas} (or equivalently we use the term \emph{domains} ${D}$ that are frequently used in the transfer learning community). That is, based on serving base stations in MR samples, the MR samples having the same serving base stations belong to the same domains. For every domain, we design a \emph{relative coordinate space} for all MR samples within the domain. We use Figure \ref{fig:relative_transfer} as an example to represent the relative coordinates. In this figure, we assume that those MR samples belonging to the same domain $D$ (a.k.a having the same serving base station $BS$) are all within a circle and $BS$ is the center. The radius $R$ of this circle is equal to the maximal distance between the positions of $BS$ and MR samples. Given this center $BS$ in the coordination space, we convert the original GPS coordinate $(x_0,y_0)$ of $BS$ into a relative one $(0, 0)$. For a MR sample $s\in D$ with the GPS coordinate $(x+x_0, y+y_0)$, its relative coordinate becomes $(x, y)$. In this way, we can compute the relative coordinates of all MR samples by referring $BS$ as the center of this coordination space.

Until now, we show the key point of the relative coordination space as follows. Let us consider another domain $D'$, where the GPS position of the serving base station (i.e., the center) belonging to $D'$ is $(x_1,y_1)$. For one MR sample $s'\in D'$ with the GPS coordinate $(x+x_1, y+y_1)$, its relative coordinate becomes $(x,y)$. Here, though the two samples $s\in D$ and $s'\in D'$ are originally with different GPS coordinates $(x+x_0, y+y_0)$ and $(x+x_1, y+y_1)$, they now share the exactly same relative coordinates $(x, y)$ under their own domains. In this way, we can perform the transfer from ${D}$ to ${D}'$. That is, {when both Telco signal strength in MR samples (a.k.a MR features) and relative position (labels) refer to serving base stations, the transfer across domains becomes possible}. Moreover, for a large amount of MR samples across an entire area, we can group the MR samples by their serving base stations to build the associated domains and relative coordination space. After the big area is divided into small domains, for each domain (and the associated serving base station), we can learn an individual mapping from MR samples within this domain to their relative positions. The key is that the mapping is adaptively learned by the data-driven fashion, even if the transmitting power, Telco signal coverage, and bandwidth of serving base stations are unavailable. Thus, even for two base stations (with various transmitting power, Telco signal coverage, and bandwidth) located at the exactly same locations, we could establish two corresponding mappings from the MR samples generated by an individual base station to the associated MR positions.

Note that the relative coordinate space above requires the GPS coordinate of serving base stations. Telco operators can easily obtain the GPS coordinates of base stations because base stations are deployed by Telco operators themselves.

2) \textbf{Representation of $F_d()$}: For a certain MR sample $s$, we convert MR features $F_d(s)$, such as \emph{RNCID} and \emph{CellID} of a neighboring base station,
into meaningful IDs which are independent upon the associated domains. Specifically, depending upon all the neighbouring base stations appearing inside the MR samples within a domain, we determine a rectangle area covered by these neighboring base stations. As shown in Figure \ref{fig:relative_transfer}, the width (resp. height) of the rectangular is equal to $2\times \max(|x_{bs_i}|)$ (resp. $2\times \max(|y_{bs_i}|)$), where we have $\max(|x_{bs_i}|) = x_{bs_1}$ and $\max(|y_{bs_i}|) = y_{bs_5}$. Then, we evenly divide the rectangle into $g \times g$ small grids (we have $g=10$ in this figure). In this way, each neighboring base station is located within a certain grid and we replace its \emph{RNCID} and \emph{CellID} by the associated grid IDs \emph{Grid\_ID}$_x$ and \emph{Grid\_ID}$_y$. For example, we represent the two base stations $bs_1$ and $bs_5$ by the grid IDs $(1, 9)$ and $(8, 1)$, respectively. The representation of $F_d(s)$ above offers the following advantage: the grid IDs are now independent upon domains and $F_d(s)\simeq F_d(s')$ holds for two MR samples $s\in D$ and $s'\in D'$.

\subsection{System Overview}
\begin{figure}[th]
\begin{center}					
\centerline{\includegraphics[height=4cm]{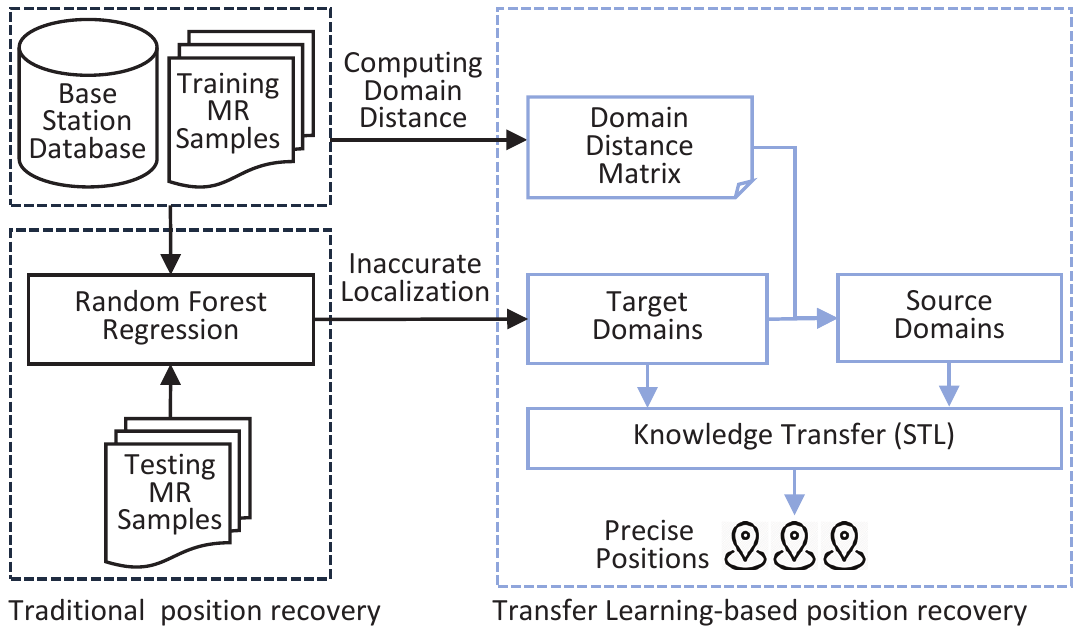}}\vspace{-2ex}
\caption{System Overview} \label{fig:transfer}\vspace{-4ex}
\end{center}
\end{figure}

Following the general idea above, we introduce three following components of \textsf{TLoc} (see Figure \ref{fig:transfer}): a traditional position recovery model (e.g., a Random Forest-based regression model), a matrix to maintain the pairwise domain similarity, and the transfer learning component for those target domains suffering from inaccurate position accuracy (caused by insufficient position labels for MR training samples).

Let us consider the following scenario in a big area, where the geo-tagged MR samples are distributed unevenly across this area.
To this end, we follow Section \ref{sec:relative} to divide the entire big area into multiple smaller areas (a.k.a \emph{domains}) and represent MR samples and associated positions under the associated relative coordinate space. Among the divided domains, due to the uneven distribution of geo-tagged MR samples, some of the domains could be with sufficient samples, and a regression-based position recovery model thus works very well. Yet other domains may contain scarce geo-tagged MR samples, the trained position recovery model usually suffers from poor localization accuracy \cite{DBLP:conf/mdm/HuangRZLYZY17}.

To this end, \textsf{TLoc} adapts the recent transfer learning scheme STL \cite{DBLP:journals/pami/SegevHMCE17} to improve the localization accuracy in the domains suffering from poor prediction precision, e.g., with a median error higher than a given threshold. We treat such domains as \emph{target domains}. Based on the developed distance metric (Section \ref{sec:source}), we choose those top-$k$ domains that \emph{1)} are most similar to a target domain and \emph{2)} are with low localization errors. Such top-$k$ domains are called the \emph{source domains} of the target one. Finally, we \emph{transfer} the recovery models from the top-$k$ source domains to the target one using an adapted STL technique (Section \ref{sec:transfer}).

\section{Domain Distance}\label{sec:source}
Since the position recovery model essentially maintains the mapping from MR features, e.g., $F_i(s) $ and $F_d(s)$, to MR positions $L(s)$, we thus define the domain distance by two parts: \emph{1}) the distance in terms of MR features and \emph{2}) the distance in terms of MR positions $L(s)$. In this section, we first give the detail for each of the two parts and next give the domain distance by integrating the two parts.

\subsection{MR Feature Distance}
To measure the similarity of MR features between two domains, the distance metric takes into account three following aspects: \emph{1}) the general approach to compute the distance of those MR features $F_i(s)$ involving Telco signal strength, \emph{2}) the distance by introducing the weight of up to seven base stations, and \emph{3}) the overall distance involving the refinement of three specific signal strengths (\emph{RSSI}, \emph{AsuLevel} and \emph{SignalLevel}).

\subsubsection{Distance of Telco Signal Strength}
\begin{figure}[th]\hspace{-2ex}
		\begin{tabular}{c c}
			\begin{minipage}[t]{0.5\linewidth}
                \includegraphics[width=3.0cm]{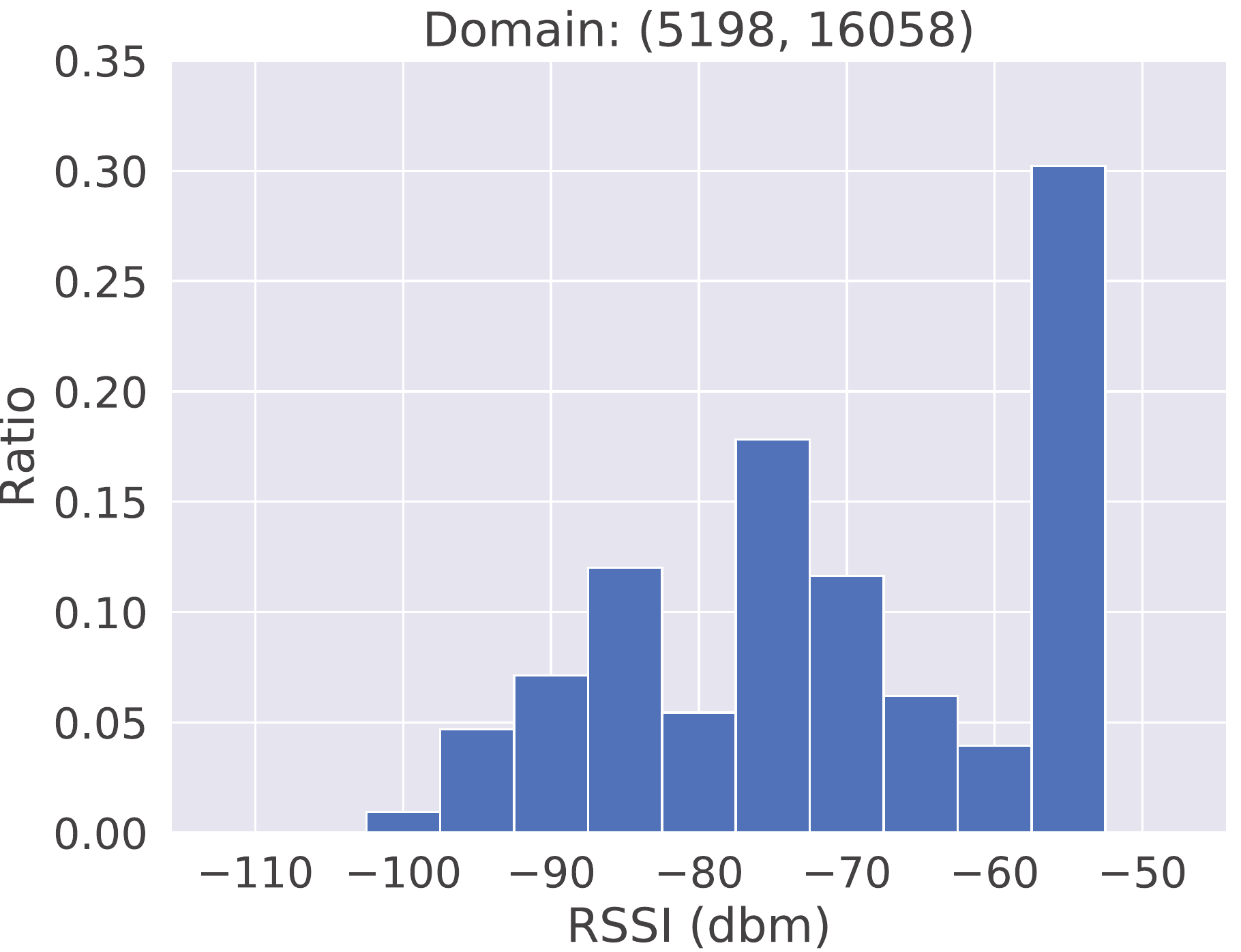}
			\end{minipage}
			&
			\begin{minipage}[t]{0.5\linewidth}
                \includegraphics[width=3.0cm]{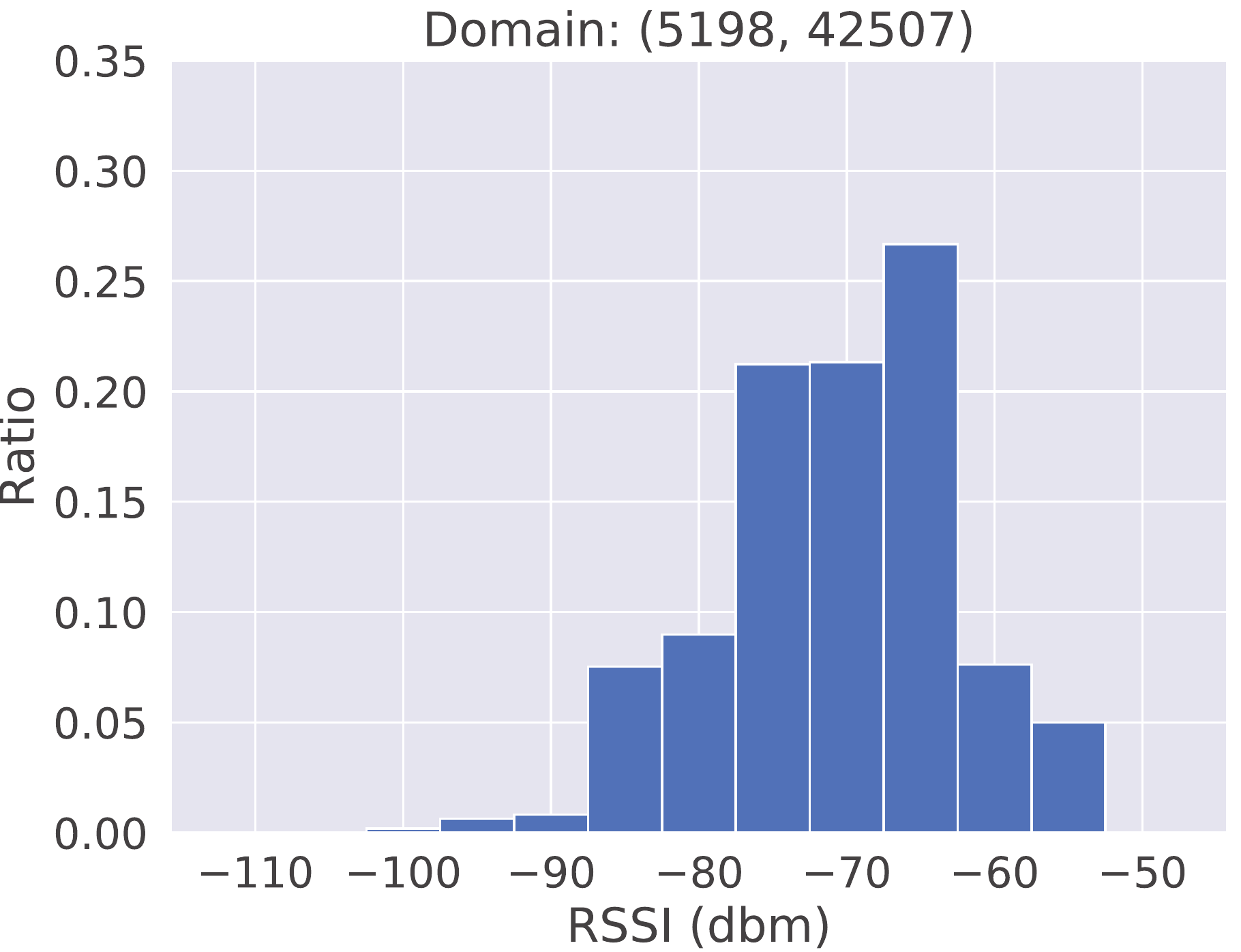}
			\end{minipage}
		\end{tabular}\vspace{-2ex}
		\caption{Histograms of Two Domains in terms of RSSI from Serving Base Station (where 5198 indicates RNC ID and 16058/42507 indicates Cell ID)}\label{exp:dist1a}\hspace{-2ex}
\end{figure}

Firstly, to compute the distance of Telco signal strength between two domains, we exploit a histogram structure to capture the overall distribution of Telco signal strength in a certain domain, and next compute the histogram distance. Figure \ref{exp:dist1a} plots the histograms of two example domains to capture the distribution of RSSI from serving base stations. The $x$-axis is the RSSI value and $y$-axis indicates the ratio of the MR samples having RSSI values falling inside a RSSI interval against total MR samples in the domain. To compute the histogram distance, we choose three frequently used metrics: probabilistic likelihood \cite{DBLP:conf/mobisys/YoussefA05, IbrahimY12}, Kullback-Leibler Divergence \cite{DBLP:journals/jlbs/MirowskiWSPMHH12}, and $p$-norm distance \cite{DBLP:journals/imwut/WuXYLY17}. Among the three metrics, we empirically find that the $p$-norm distance with $p=3$ leads to the best result. Formally, for a domain $D$ (resp. $D'$), we denote by $h_{D,j}$ (resp. $h_{D',j}$) the MR sample rate in $y$-axis for the $j$-th RSSI interval in $x$-axis. When each histogram contains $r$ RSSI intervals, we compute the histogram distance between $D$ and $D'$ by

\begin{equation}\label{eq:hist}\scriptsize
dis_{hist}(D,D')=(\sum_{j=1}^{r}(|h_{D,j}-h_{D',j}|)^p)^\frac{1}{p}
\end{equation}

\subsubsection{Weighted Distance of Telco Signal Strength}
We note that each MR sample contains up to seven base stations sorted by descending order of Telco signal strength. These stations contribute differently to the distance in Equation (\ref{eq:hist}), due to various signal strength caused by these stations.

\begin{figure}[th]
\begin{center}
	\begin{tabular}{c}
		\begin{minipage}[t]{1.0\linewidth}
            \begin{center}
				\centerline{\includegraphics[width=5cm]{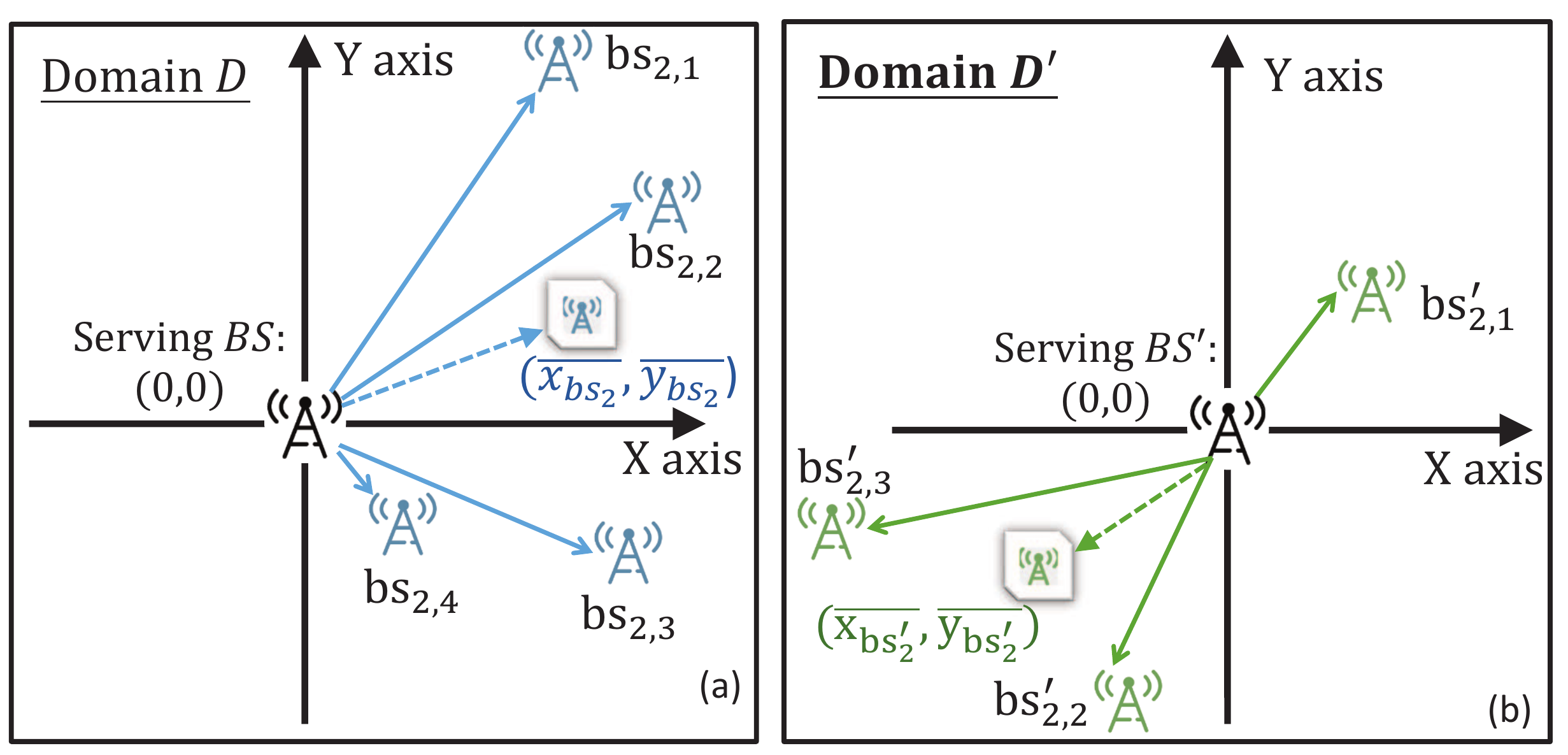}}
			\end{center}
		\end{minipage}\vspace{-2ex}
		\\
		\begin{minipage}[t]{1.0\linewidth}
		  \begin{center}
				\centerline{\includegraphics[width=5cm]{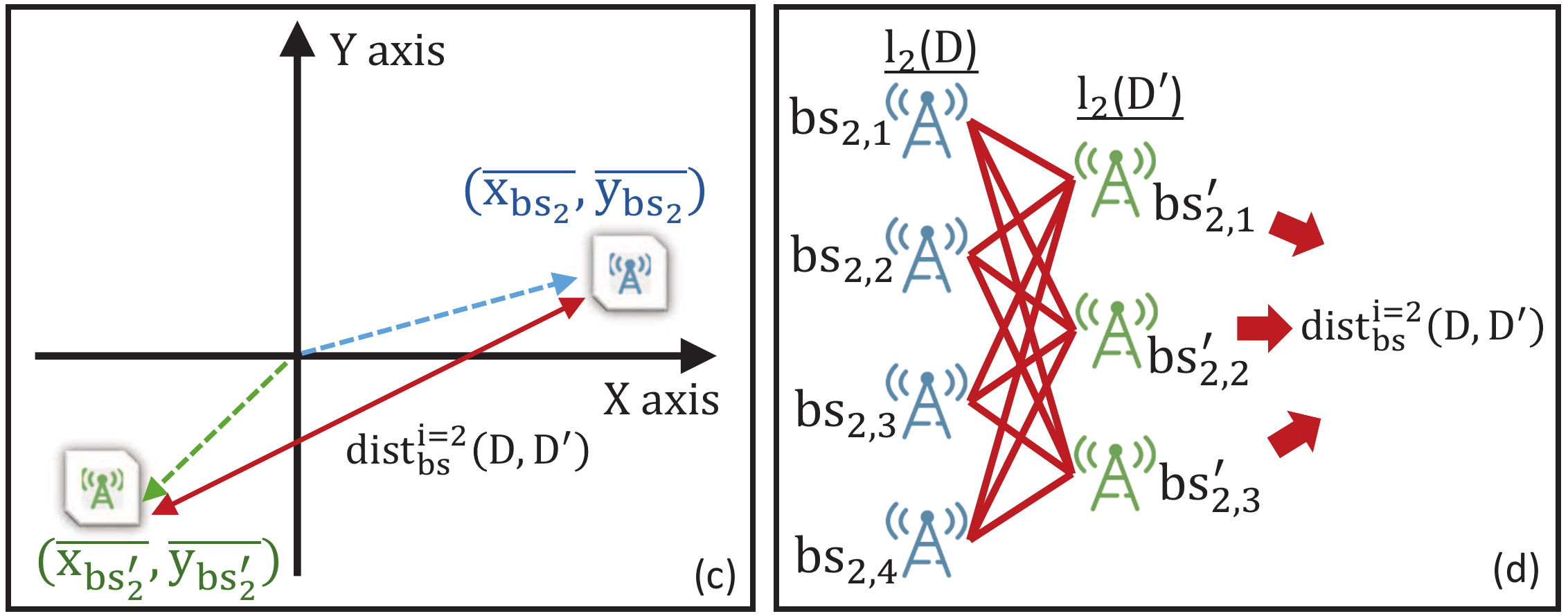}}
		   \end{center}
		\end{minipage}
	\end{tabular}\vspace{-2ex}
\caption{Distance of BS Locations between Domains} \label{fig:bs_dis}\vspace{-2ex}
\end{center}
\end{figure}

Specifically, each domain $D$ contains a set of MR samples. For all (neighboring) base stations appearing in these MR samples, we group such stations by their order index in MR samples: the 1st group contains only one serving base station with the strongest signal strength, the 2nd group contains the neighboring base stations of 2nd order index (i.e., RNCID\_2, CellID\_2) in each MR sample. In this way, we have up to seven groups of base stations. Each group contains a list of base stations, denoted by $l_i$ with $i=1,...,7$. In this way, we improve Equation (\ref{eq:hist}) by introducing a weight $w_i$ for each $l_i$.
\begin{equation}\label{eq:whist}\scriptsize
\begin{aligned}
dis_{mr}(D,D')&\!= \!\sum_{i=1}^{7}w_i \! \times \! dis_{hist}^{i}(D,D')\\
\end{aligned}
\end{equation}

In Equation (\ref{eq:whist}), $dis_{hist}^{i}(D,D')$, computed by Equation (\ref{eq:hist}), is the histogram distance for the Telco MR signal associated with the $i$-th lists $l_i(D)$ and $l_i(D')$ in two domains $D$ and $D'$, and $w_i$ is the weight of the $i$-th group.

We give the general idea of computing the weight $w_i$ as follows. Recall that Section \ref{sec:relative} transforms the neighboring base stations (identified by \emph{RNCID} and \emph{CellID}) into grid IDs. Such grid IDs approximate the positions of neighboring stations within each domain: the nearest (resp. farthest) base stations contribute to the strongest (resp. weakest) Telco signal strength. We leverage these grid IDs to compute the weight $w_i$. As shown in Figure \ref{fig:bs_dis}, we have the 2nd base station lists in two domains $D$ and $D'$, denoted by $l_2(D)$ and $l_2(D')$, which contain 4 member stations $bs_{2,1}...bs_{2,4}$ and 3 stations $bs'_{2,1}...bs'_{2,3}$, respectively. Based on the distance $dis_{bs}^{i=2}(D,D')$ of the two lists $l_2(D)$ and $l_2(D')$, , we define the normalized weight $w_i$ as follows.

\begin{equation}\label{eq:w_dis}\scriptsize
\begin{aligned}
w_i &= \frac{e^{dis_{bs}^i}}{\sum_{j=1}^{7}e^{dis_{bs}^j}} \\
\end{aligned}
\end{equation}

To compute the item $dis_{bs}^{i}$ above, as shown in Figure \ref{fig:bs_dis}(c), we exploit the average position of the 4 stations in $l_2(D)$, denoted by $(\overline{x_{bs_i}}, \overline{y_{bs_i}})$, and one of the 3 stations in $l_2(D')$, denoted by $(\overline{x_{bs'_i}}, \overline{y_{bs'_i}})$. After that, we compute the Euclidean distance between the two average positions.

\begin{equation}\label{eq:method1_dis}\scriptsize
dis_{bs}^{i}(D,D') = [(\overline{x_{bs_i}}-\overline{x_{bs'_i}})^2+(\overline{y_{bs_i}}-\overline{y_{bs'_i}})^2]^{\frac{1}{2}}
\end{equation}

Nevertheless, the average positions above might lose the geographical characteristics of base stations. Thus, as an improvement to compute the item $dis_{bs}^{i}$, as shown in Figure \ref{fig:bs_dis}(d), we first calculate the pairwise distance between the base stations in $l_i(D)$ and $l_i(D')$, and then compute the average of the pairwise distance:
\begin{equation}\label{eq:method2_dis}\scriptsize
\begin{aligned}
dis_{bs}^{i}(D,D') = \frac{\sum_{n=1}^{|l_i(D')|}\sum_{m=1}^{|l_i(D)|}ed(bs_{i,m},bs'_{i,n})}{|l_i(D)||l_i(D')|}\\
\end{aligned}
\end{equation}
In Equation (\ref{eq:method2_dis}), $ed(\cdot)$ indicates the Euclidean Distance of two base stations $bs_{i,m}$ and $bs'_{i,n}$, whose positions are represented by grid IDs under the relative coordinate space.

Note that the 4G LTE MR samples collected by Android API may miss the IDs of neighbour base stations (see Section \ref{s:problem}), and we cannot leverage the positions of base stations, required by Equation (\ref{eq:method2_dis}), to compute the weight $w_i$. To overcome this issue, we could approximate $w_i = \frac{1/i}{\sum_{j=1}^{7} 1/j}$, such that $w_i$ is inverse to the index number $i$. This approximation makes sense: the index number $i$ essentially indicates the signal strength of base stations, and the weight $w_1$ regarding to the 1st index (i.e., the serving base station) consequently contributes most to the overall distance.

\subsubsection{Overall Distance of MR Features}
Thirdly, recall that MR samples in Table \ref{tab:mr} contain three types of signal strength: \emph{RSSI}, \emph{AsuLevel} and \emph{SignalLevel}. For such signal strength, we might first follow Equation (\ref{eq:whist}) to compute three associated histogram distance such as $dis_{mr}^{rssi}(D,D')$ and next sum the three weighted distance as the overall distance. However, the sum may not provide a sensible overall measure if the three types of Telco signal strength are heavily dependent. In fact this is the case because \emph{AsuLevel} is a scaling value of \emph{RSSI}, i.e., in 2G GSM data set, $AsuLevel=(RSSI+113)/2$ \cite{Rizk2019CellinDeep}, and $dis_{mr}^{asu}$ can be treated as a linear transformation of $dis_{mr}^{rssi}$. Among the three types of signal strength, we thus take into account the independent contribution of \emph{RSSI} and \emph{SignalLevel} to compute the overall distance of MR features.

\begin{figure}[th]
		\begin{tabular}{c c}
			\begin{minipage}[t]{0.5\linewidth}
                \includegraphics[width=4.0cm]{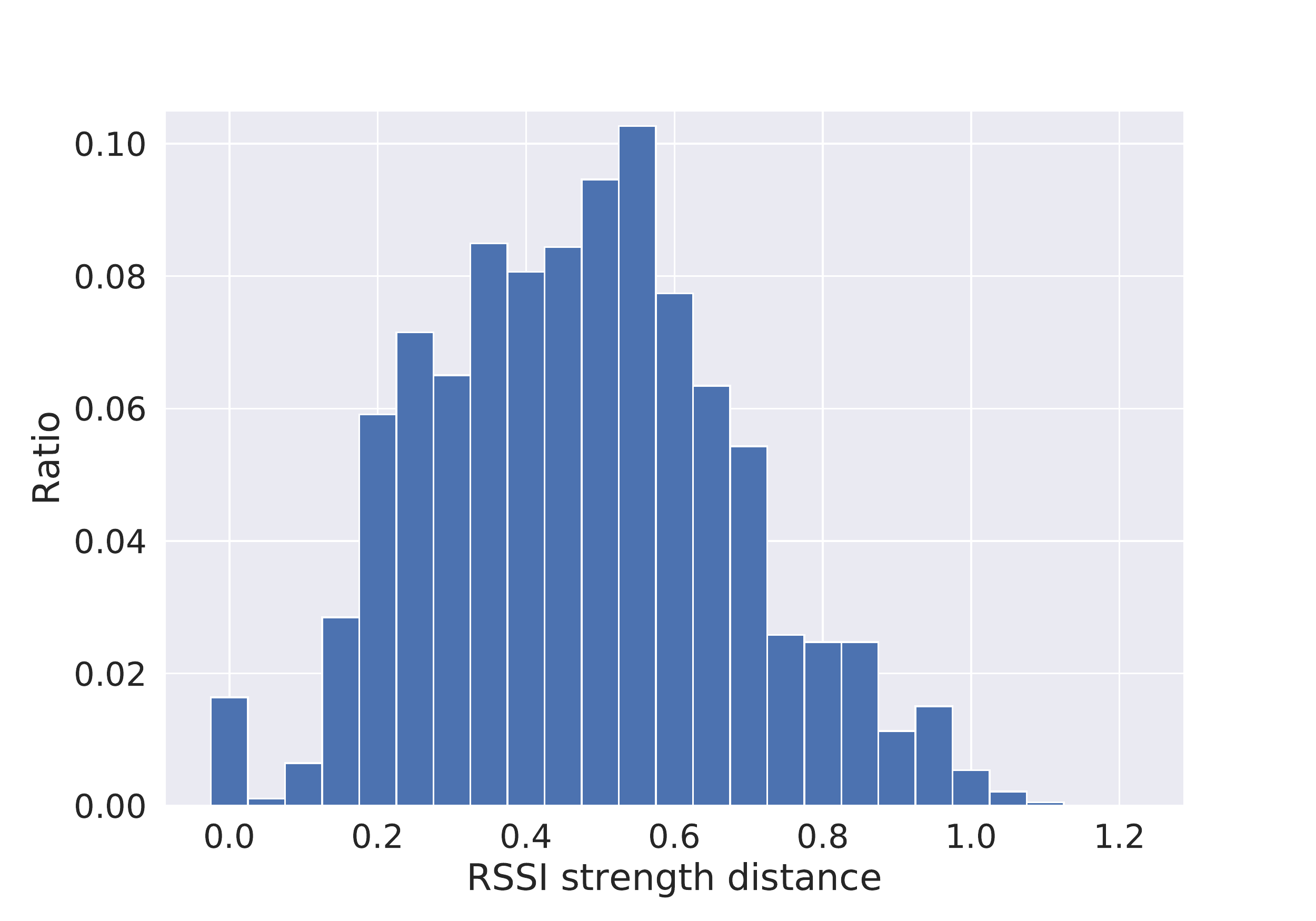}
			\end{minipage}
            &
            \begin{minipage}[t]{0.5\linewidth}
					\includegraphics[width=4.0cm]{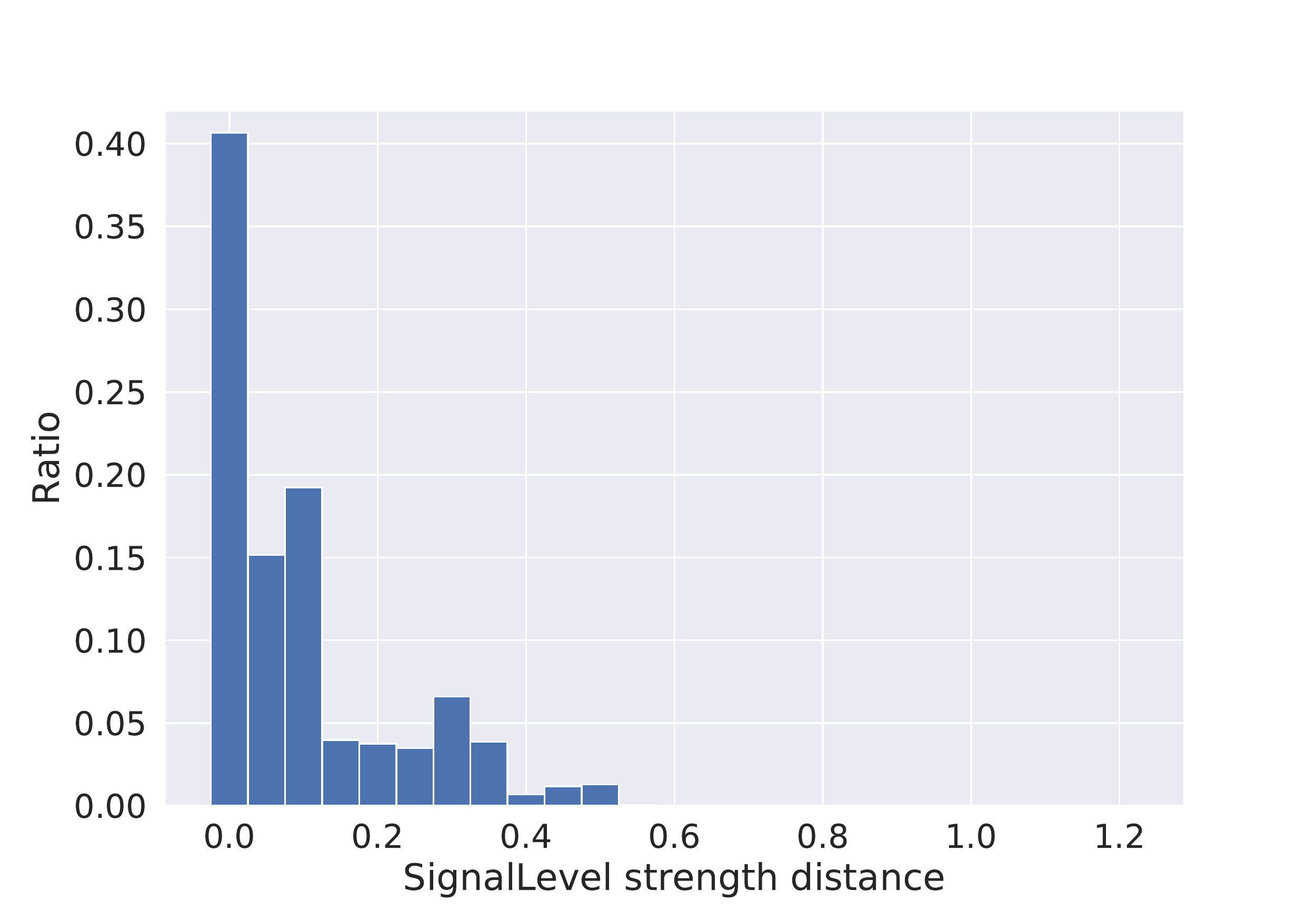}
			\end{minipage}
		\end{tabular}\vspace{-2ex}
  \caption{Distribution of RSSI (left) and SignalLevel (right) Histogram Distance between Pairwise Domains.}\vspace{-2ex}
  \label{fig:heatmap}
\end{figure}

In terms of \emph{SignalLevel}, we follow Equation (\ref{eq:whist}) to compute its histogram distance $dis_{mr}^{sig}(D,D')$. As shown in Figure \ref{fig:heatmap}, the distribution of $dis_{mr}^{sig}(D,D')$ in our datasets significantly differs from the one of $dis_{mr}^{rssi}(D,D')$: around 40\% \emph{SignalLevel} distance values are 0.0 and more than 85\% (resp. 95\%) are smaller than 0.1 (resp. 0.3). The numbers indicate that the majority of \emph{SignalLevel} feature values in the datasets are zeros and the output and input signals are equal (see Section \ref{s:problem} for the meaning of \emph{SignalLevel}). Thus, we could assign a small weight for \emph{SignalLevel} and among the overall distance, the distance of \emph{SignalLevel} contributes less than the one of \emph{RSSI}. Since \emph{RSSI} distance plays a key role in the overall distance, we use the \emph{average Pearson coefficient} $c$ between \emph{RSSI} and \emph{SignalLevel} as the weight of \emph{SignalLevel}.


Based on the intuition above, we compute the overall distance of MR features as follows.

\begin{equation}\label{eq:signal}\scriptsize
\begin{aligned}
dis_{mr}(D,D') &=\frac{1\times dis_{mr}^{rssi}(D,D')+c\times dis_{mr}^{sig}(D,D')}{1+c} \\
dis_{mr}^{rssi} &= \sum_{i=1}^{7}w_i \! \times dis^{i}_{hist\_rssi}(D,D')\\
dis_{mr}^{sig} &= \sum_{i=1}^{7}w_i \! \times dis^{i}_{hist\_sig}(D,D')
\end{aligned}
\end{equation}

In the equation above, $dis_{mr}^{rssi}(D,D')$ (resp. $dis_{mr}^{sig}(D,D')$) is the weighted histogram distance between $D$ and $D'$ for \emph{RSSI} (resp. \emph{SignalLevel}) using Equation (\ref{eq:whist}), and $c$ is the average Pearson coefficient between \emph{RSSI} and \emph{SignalLevel}.


\subsection{Relative Position Distance}
Besides MR features, we also compute the distance of MR positions (labels) between two domains. Since we have represented MR positions by relative ones, we compute the distance by relative positions. In addition, instead of using discrete MR positions, we connect such positions into moving trajectories.

For one mobile device (identified by IMSI), we have a series of relative positions corresponding to the neighbouring MR samples sorted by the MR time stamp. In the case where the timestamp gap between any two neighbouring MR samples exceeds a threshold (e.g., 60 minutes), we divide the MR series into multiple short ones. A short MR series then becomes an associated moving trajectory. The trajectories are useful for understanding the overall spatio-temporal mobility patterns of mobile devices. Thus, we compute the distance of the trajectories, instead of MR positions, between two domains.

Given two trajectories $T$ and $T'$, we compute the Frechet distance  \cite{DBLP:books/lib/BergCKO08}: $dis(T,T')=min [max_{t\in T, t'\in T'}dis(t,t')]$, where $t$ and $t'$ indicate the sample points in trajectories $T$ and $T'$, respectively. If an Euclidean distance is used to compute $dis(t,t')$, then the sub-item $max_{t\in T, t'\in T'}dis[t,t']$ computes the \emph{maximum distance}, and the item $min [max_{t\in T, t'\in T'}dis(t,t')]$ finds the minimal one among the maximum distance.

\begin{figure}[th]
    \begin{center}
    \begin{tabular}{c c}
			\begin{minipage}[t]{0.48\linewidth}
                \includegraphics[height=1.2in]{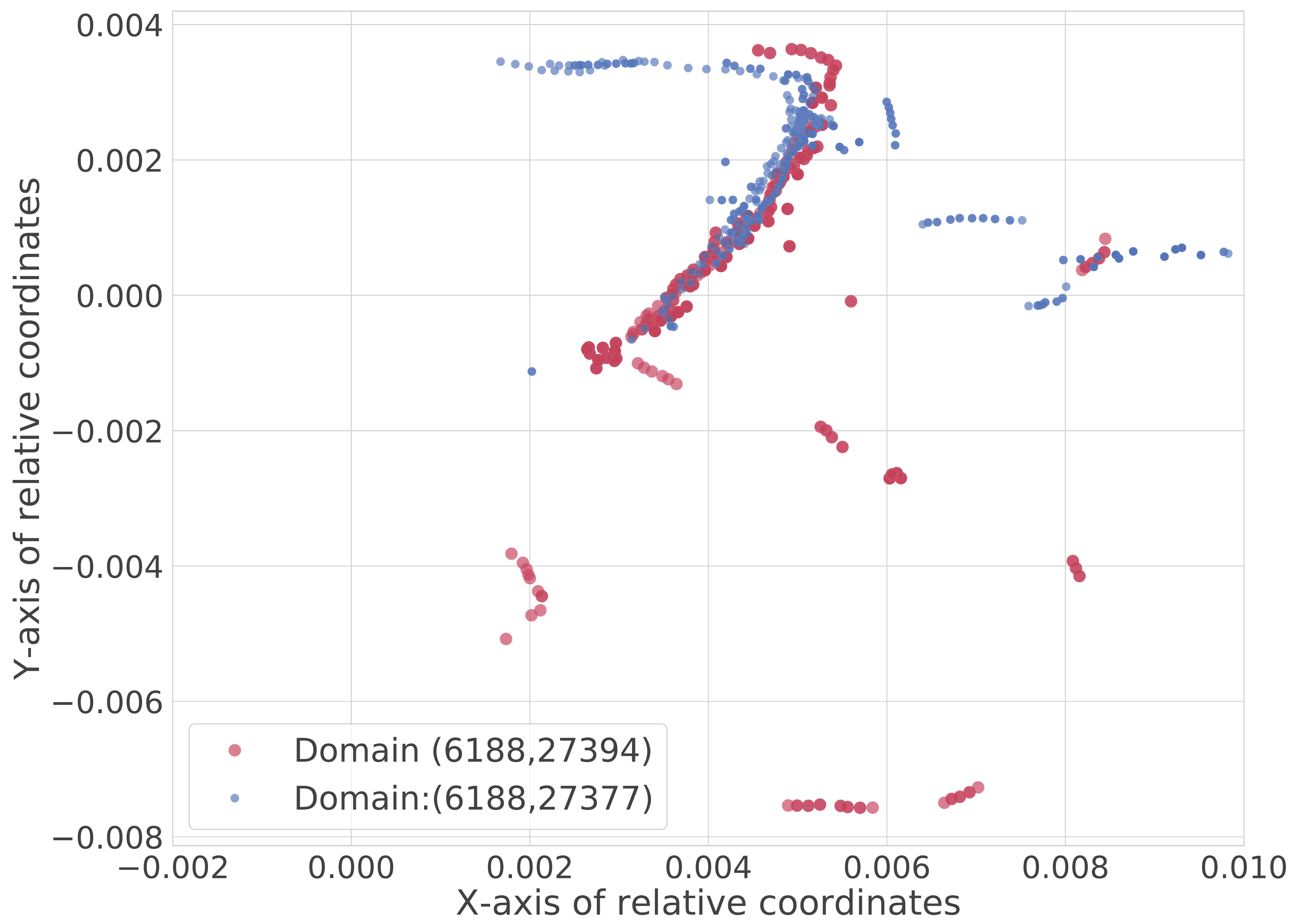}
			\end{minipage}
			&
			\begin{minipage}[t]{0.48\linewidth}
                \includegraphics[height=1.2in]{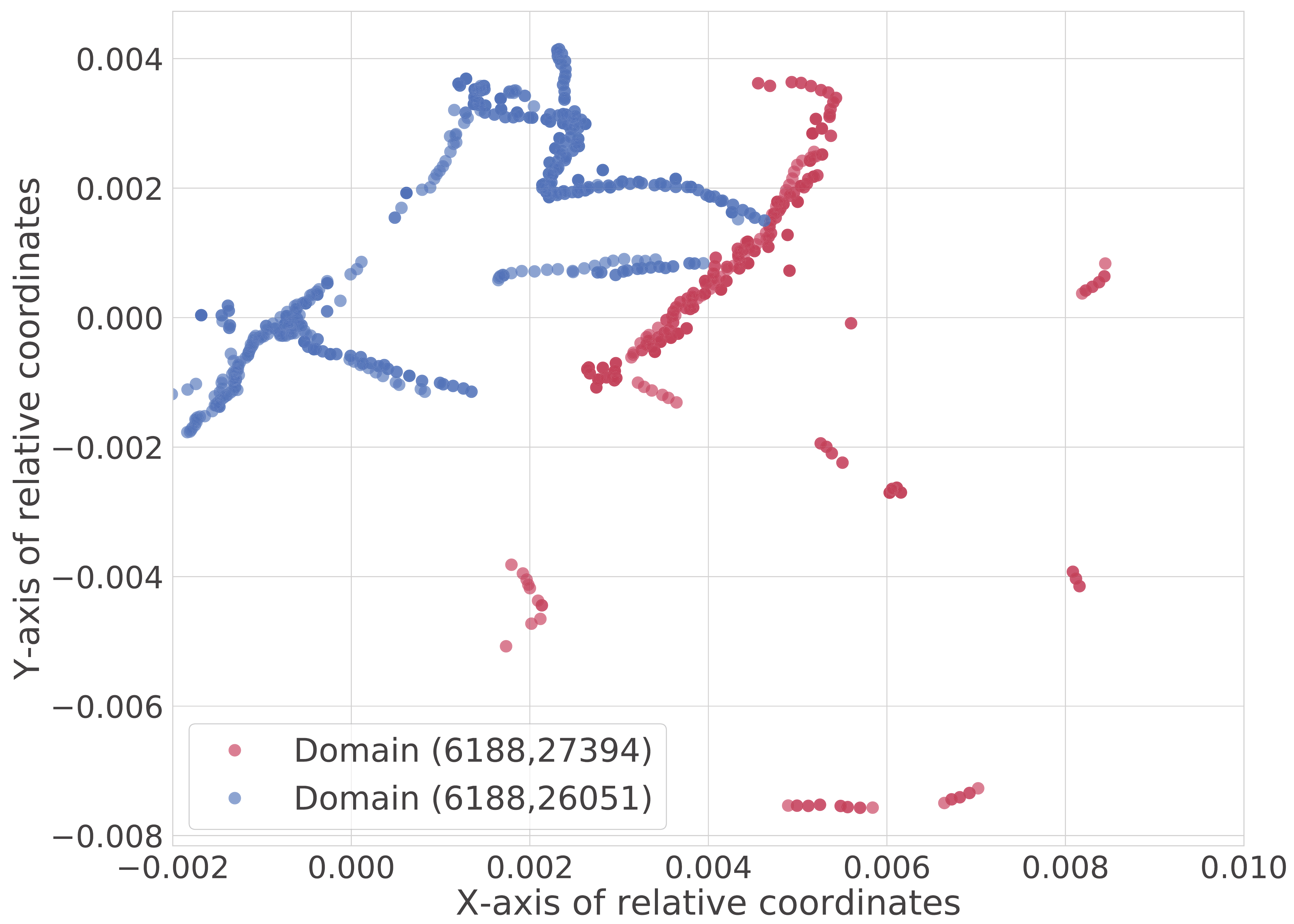}
			\end{minipage}
		\end{tabular}\vspace{-2ex}
    \end{center}
	\caption{Trajectory distance of the left figure is smaller than the right one, where 6188 is RNCID and 27394 is CellID.}		\label{exp:dist1b}
\end{figure}

In addition, each domain may contain multiple trajectories. Thus, we compute the average of the sum of pairwise trajectory distance:
\begin{equation}\scriptsize
dis_{pos}(D,D')= \frac{\sum_{T\in D, T'\in D'}{dis(T,T')}}{|D|\times|D'|}
\end{equation}

where $|D|$ and $|D'|$ indicate the trajectory count in domains $D$ and $D'$, respectively. $dis_{pos}(D,D')$ indicates the average distance between any two trajectories in $D$ and $D'$. As shown in Figure \ref{exp:dist1b}, the trajectory distance between two domains (6188, 27394) and (6188, 27377) is smaller than the one between two domain (6188, 27394) and (6188, 26051).

\subsection{Source Domain Selection by Domain Distance}
We now integrate the two distance of MR features and positions above to define the overall domain distance.
\begin{eqnarray}\label{eq:overalldist}\small
dist(D,D')=w_{mr} \times  dis_{mr}+w_{pos} \times  dis_{pos}
\end{eqnarray}

In the equation above, the weights $w_{mr}$ and $w_{pos}$ with $0\leq w_{mr}, w_{pos}\leq 1.0$ and $w_{mr}+ w_{pos} = 1.0$ measure the importance of $dis_{mr}$ and $dis_{pos}$, respectively. By default we set $w_{mr}=w_{pos}=0.5$. Our evaluation will show that such parameters can be effectively tuned according to the amount of labelled samples in source and target domains.

\begin{figure}[th]
	\begin{center}
		\begin{tabular}{c c}
			\begin{minipage}[t]{0.5\linewidth}
				\begin{center}
					\centerline{\includegraphics[height=1.4in]{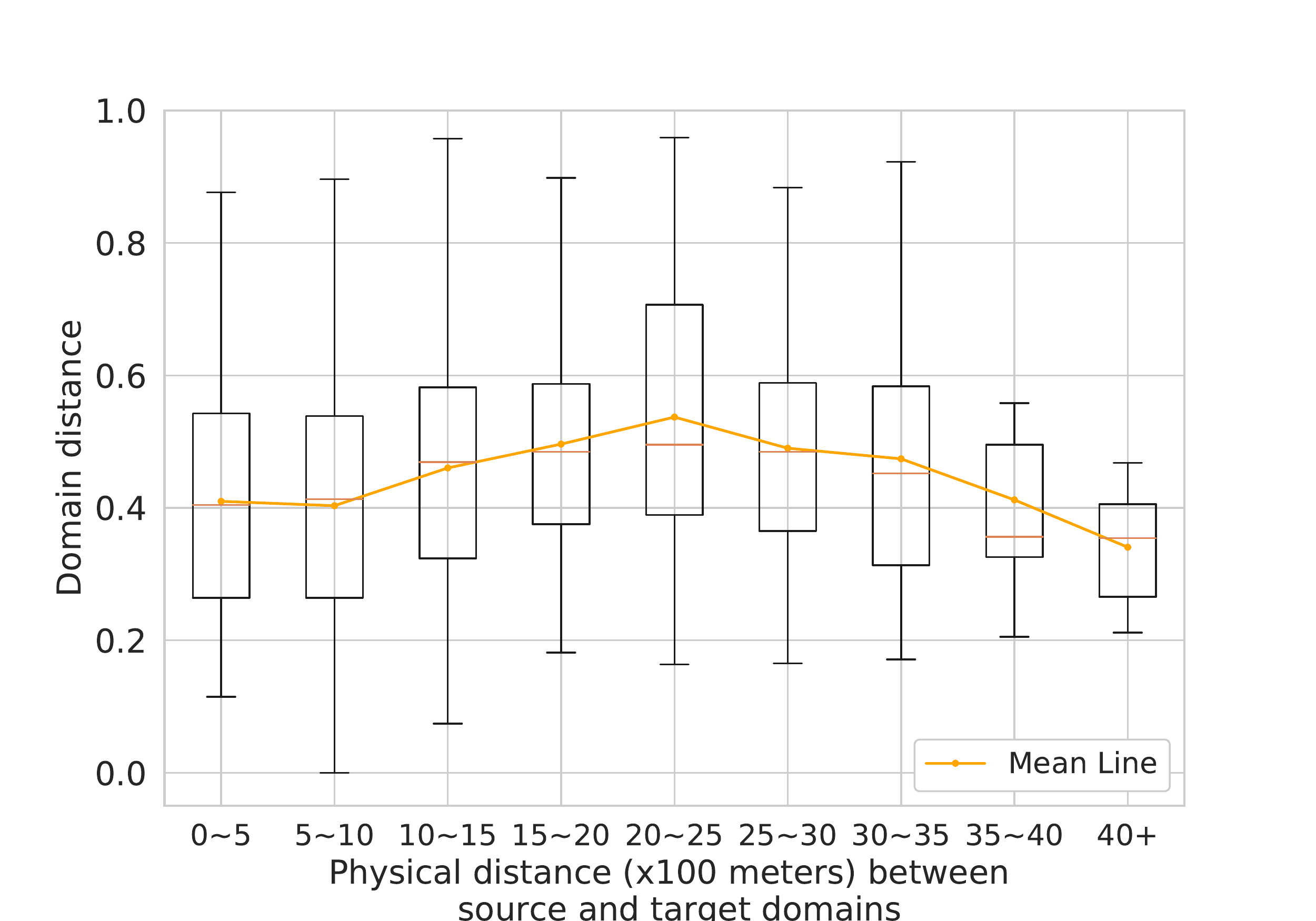}}
				\end{center}
			\end{minipage}
			&
			\begin{minipage}[t]{0.5\linewidth}
				\begin{center}
					\centerline{\includegraphics[height=1.4in]{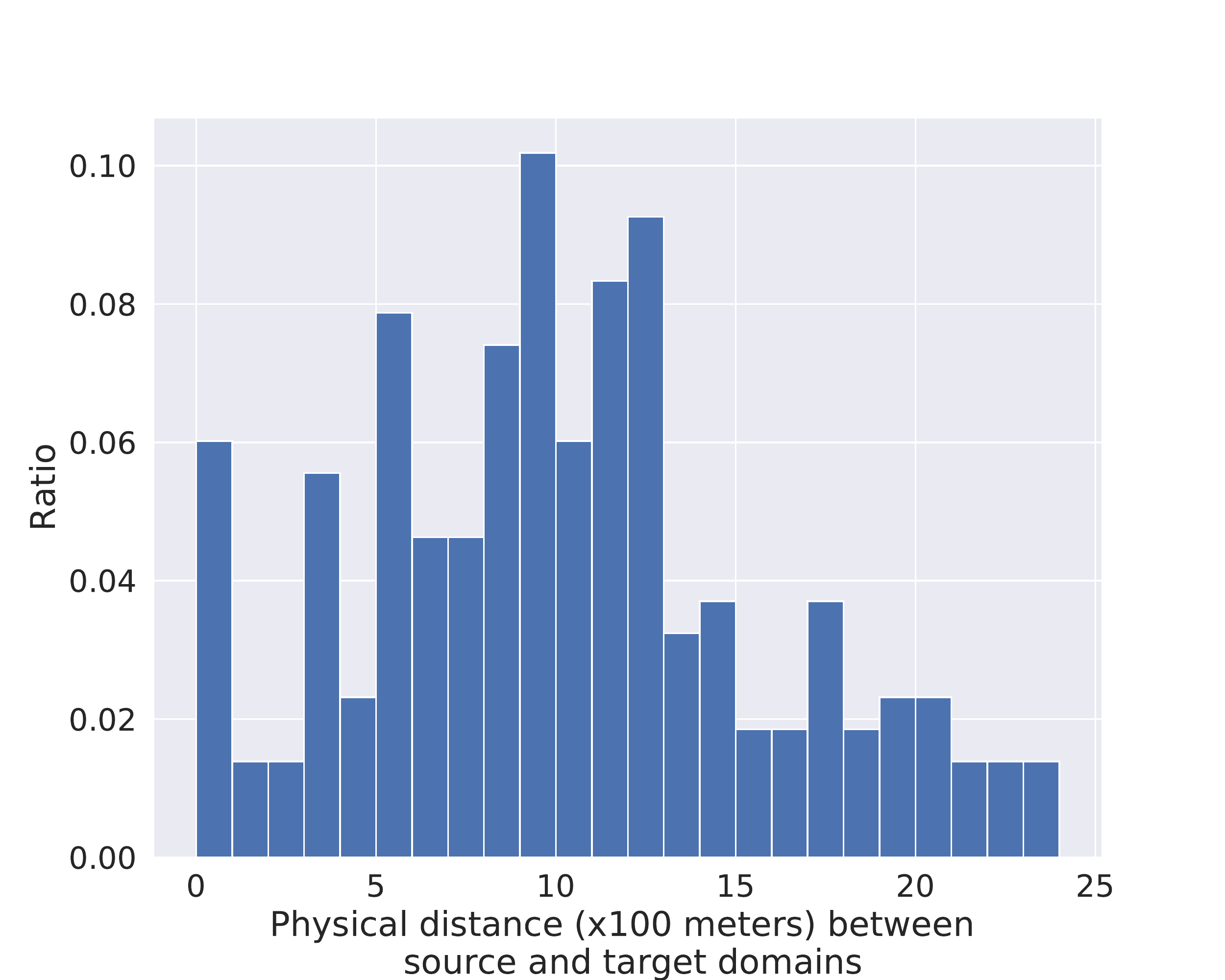}}\vspace{-3ex}
				\end{center}
			\end{minipage}
		\end{tabular}
	\end{center}\vspace{-4ex}
	\caption{Domain Selection (Jiading 2G). Left: Domain Distance; Right: Top-$k$ source domains}\label{fig:dis_range}
\end{figure}

Given the defined metric above, we are interested in how the similar domains are also physically close. To this end, for our Jiading 2G GSM data set, we plot Figure \ref{fig:dis_range} (left) to give the average domain distance under various physical distance between domains. The $x-$axis indicates the interval of the physical distance, and the $y$-axis is the average domain distance within the interval. This figure indicates that two physically closer domains, e.g., the physical distance is smaller than 2.5 km, are more similar. Moreover, two domains, though rather far away, still have chance to be similar.

Next, Figure \ref{fig:dis_range} (right) plots the physical distance between top $k=3$ source domains and a target one, where $x-$axis is the interval of physical distance between source and target domains, and $y-$axis is the rate of source domains. We find that the distance between most source and target domains is smaller than 2.5 km, consistent with Figure \ref{fig:dis_range}.

From  Figure \ref{fig:dis_range}, we find that the needed source domains for a target one are physically close. In addition, some far-away domains are useful for a target one (In Section \ref{s:evaluation}, we select source domains across different areas which leads to the best transferring results). Thus, we compute the pairwise domain distance for the top-$k$ most similar source domains for the target ones. Nevertheless, when the count of divided domains is a large number, the pairwise domain distance involves non-trivial computing overhead. Thus, for higher efficiency, we apply the locality sensitive hash (LSH) technique \cite{DBLP:conf/stoc/IndykM98} to approximately find the top-$k$ source domains for a target one. Our experiment will investigate the trade-off between approximation precision and computation efficiency.

\section{Structure Transfer in Random Forests}\label{sec:transfer}

\begin{figure}[th]
\begin{center}					
\centerline{\includegraphics[height=3.2cm]{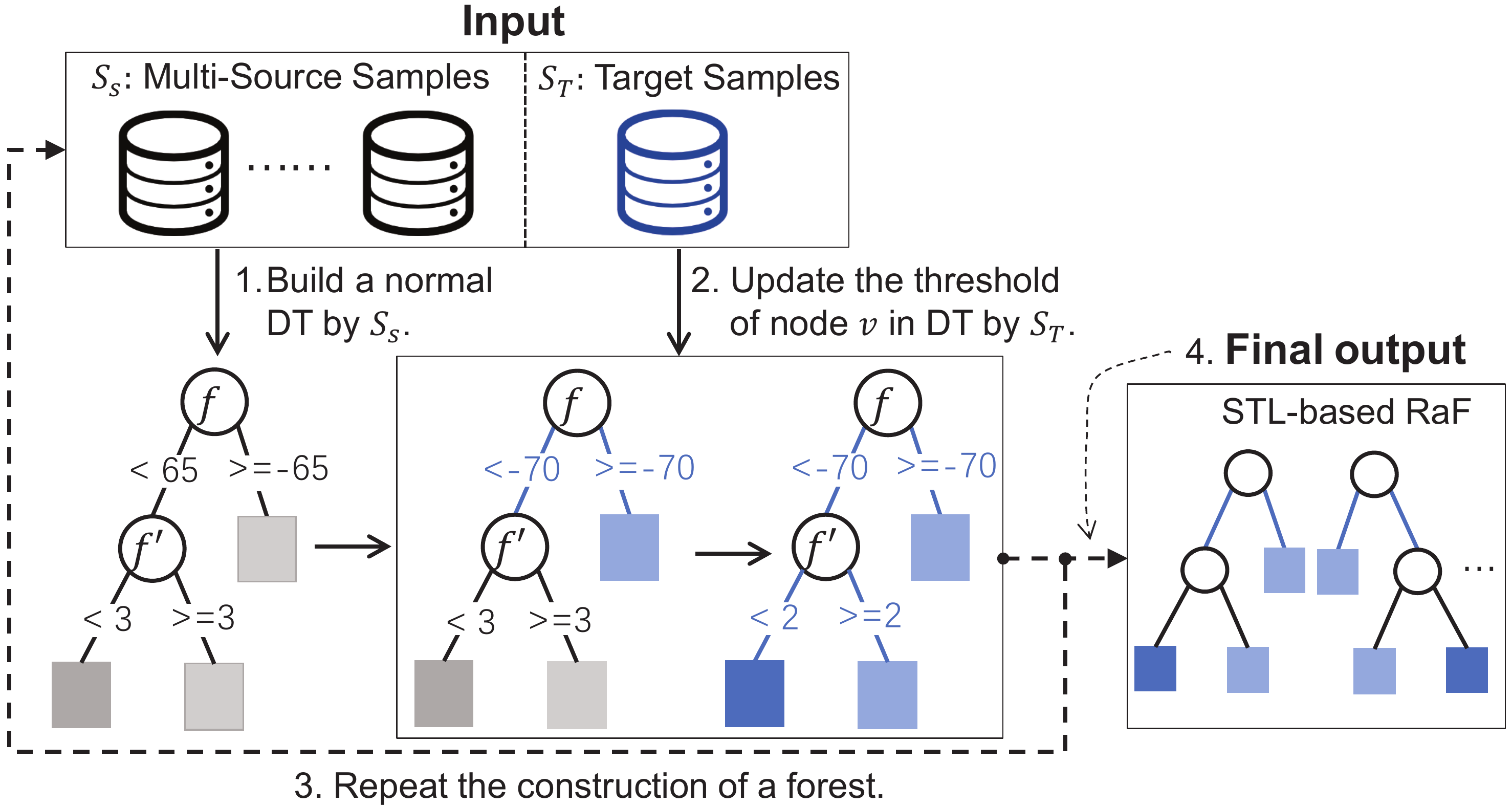}}\vspace{-2ex}
\caption{Details of Structure Transfer in Random Forest.} \label{fig:STL}
\end{center}\vspace{-2ex}
\end{figure}

In this section, we give the detail of the proposed transfer learning framework on a Random Forest (RaF) regression model. We consider the labelled MR samples (denoted by $\mathcal{S}_T$) in a target domain and those (denoted by $\mathcal{S}_{\mathcal{S}}$) in the top-$k$ source domains. A simply way is to mix the data samples from $\mathcal{S}_T$ and $\mathcal{S}_S$, and then apply a classic RaF algorithm \cite{breiman01}. However, this approach cannot  differentiate source domains from the target one, and thus does not work very well.

To solve the issue above, we adapt the recent structure transfer learning (STL)  \cite{DBLP:journals/pami/SegevHMCE17} (its general idea refers to Section \ref{s:problem}) to solve a regression model that differs from the classification problem in the original STL work \cite{DBLP:journals/pami/SegevHMCE17}. Figure \ref{fig:STL} gives the work flow of STL. The input of STL is the labelled MR samples in $\mathcal{S}_T$ (target domain) and $\mathcal{S}_S$ (multi-source domains), and the output is a transferred random forest model which is adaptive to the target domain. Specifically, we first use the data samples from $\mathcal{S}_S$ (i.e., those $k$ selected source domains) to determine the feature $f$ which can perform the split at each node $v$ in a certain decision tree (DT). Next, we re-calculate the node split thresholds $f_\phi$ by using only the data from $\mathcal{S}_T$. For example, still in Figure \ref{fig:STL}, the original threshold of feature $v$ at node $v$ is -65 computed by source samples $\mathcal{S}_S$ only. Then the threshold is optimized to -70 by the target samples $\mathcal{S}_T$. In this way, STL works top-down to select a new threshold for each node, and finally generates a random forest with transferred DTs.

Ideally, a desirable threshold yields high similarity between the distributions transferred from source domains to the target one. The purpose is that the threshold is adaptive to the target domain. Meanwhile, this similarity is restricted to ``informative" thresholds where, for any sufficiently small $\epsilon > 0$, the information gain (IG) of threshold $x$ is larger than the IG of any other $x' \in (x-\epsilon,x+\epsilon)$ in the $\epsilon$-neighbourhood of $x$. It means that the thresholds are local maximums of IG. We thus formulate the threshold selection as an optimization problem.

\begin{equation}\label{eq:6}\small
\begin{aligned}\small
\operatorname{Max}_x &DG
\left(Q^t_v,f,x,P_{v,left}(f),P_{v,right}(f)\right)\\
s.t.\quad & x \! \in \! \mathbb{R}, \forall x'  \! \in \! (x-\epsilon,x+\epsilon):\\
          &
          -\left[h(x)-mean(y)\right]^2  \leq  -\left[h(x')-mean(y)\right]^2
\end{aligned}
\end{equation}
In the equation above, $Q^t_v$ denotes the samples of target task $t$ at node $v$, $h(x)$ (resp. $h(x')$) is the prediction of $x$ (resp. $x'$), $mean(y)$ is the mean of label $y$, and $DG$ is the Jensen-Shannon divergence gain defined on Kullback-Leibler divergence and mean distribution. In addition, $P_{v,left}(f)$ and  $P_{v,right}(f)$ indicate that label distribution of two subsets (left and right) split on the feature $f$ at node $v$. The optimization problem in Equation (\ref{eq:6}) uses \emph{DG} to quantify distributional similarity and information gain criterion computed by $-\left[h(x)-mean(y)\right]^2$ to measure a threshold's informative value. Thus, the solution of this optimization problem maximizes the defined similarity $DG$ to make sure that the optimal decision threshold $f_\phi$ is adaptive enough to the target domain, thus leading to a better decision threshold $f_\phi$.

\section{Evaluation}\label{s:evaluation}

\begin{table}
\scriptsize
\centering
\begin{tabular}{|l|l|l|l|l|}
\hline
	&\textbf{\emph{Jiading}}: 2/4G &\textbf{\emph{Siping}}: 2/4G &\textbf{\emph{Xuhui}}: 2/4G  &\textbf{\emph{Urumqi}}: 2G\\
\hline
\hline
\# of samples&15954/10372 & 6723/4953 & 13404/7755 &7645\\\hline
Route Len: km & 94.1/52.1 & 24.6/15.5 & 26.4/12.7 &17.3\\\hline
\# of samples/sec& 2$\thicksim$3 & 2$\thicksim$3  &  1  &2$\thicksim$3\\\hline
BS density& 25.85/29.43 & 27.16/34.67 & 28.18/37.12 &18.31\\\hline
\# of serving BSs & 61/44  &51/42 & 21/16 &39\\\hline
\end{tabular}
\caption{Statistics of Used Data Sets (BSs: base stations)}\label{tab:dataset}\vspace{-2ex}
\end{table}

\subsection{Datasets and Counterparts}
\textbf{Datasets}: In Table \ref{tab:dataset}, we mainly use seven data sets collected at two cities in China: Shanghai and Urumqi. The data sets in Shanghai are sampled from three areas: 1) a university campus in the sub-urban area {\emph{Jiading}}, 2) another university campus in the urban area \emph{Siping}, and 3) several main roads in the core urban area \emph{Xuhui}. The average physical distance of the three areas is around 15-37 km. In each of the three areas in Shanghai, we have two data sets containing MR records collected from 2G GSM and 4G LTE networks. The data sets in \emph{Xuhui} were sampled from backend cellular towers, and the data sets in {\emph{Jiading}} and \emph{Siping} campus were collected by our developed Android mobile app via frontend Android API. In addition, to generally validate the performance of \textsf{TLoc} in various cities, we collect a 2G GSM MR data set by our app in \emph{Urumqi}, where only 2G GSM Telco network is available. Since the \emph{Urumqi} dataset contains a relatively small quantity of MR samples, we by default evaluate \textsf{TLoc} on the Shanghai data sets without special mention.

For the mobile phones installed with our app, mobile users holding these mobile phones moved around the road network inside the campus. The app then collected MR samples and GPS coordinates. Specifically, when collecting MR samples from a Telco network, the mobile app switches on GPS receivers and records the current GPS coordinates. The collected GPS coordinates are used as the location ground truth. Note that the GPS coordinates collected by mobile phones may contain noise. We thus employ the data cleaning techniques including map-matching to mitigate the effect of noise \cite{ZhengCWY14a}.

\textbf{Counterparts}: We compare \textsf{TLoc} against four previous works and two variants of \textsf{TLoc} (see Table \ref{tab:methods}).

\begin{small}
\begin{table}
\scriptsize
\centering
\begin{tabular}{|l|l|l|}
\hline
\textbf{Counterpart} & \textbf{Description} & \textbf{Source Selection}\\\hline\hline
NBL \cite{MargoliesBBDJUV17} & Recent fingerprinting method& No transfer\\
CellSense \cite{IbrahimY12}  & Classical fingerprinting method & No transfer \\
DeepLoc \cite{DBLP:conf/gis/ShokryTY18}& Recent deep neural network method  & No transfer \\
Non-Transfer \cite{ZhuLYZZGDRZ16} & Random Forest regression & No transfer\\\hline
MTL \cite{DBLP:journals/ieicet/SimmAS14} & Multi-task learning in Random Forest& No src selection\\
SVR-Transfer \cite{DBLP:conf/aaai/ZhengPYP08}& Transfer Learning in SVR & No src selection\\\hline
\textsf{TLoc} & Our approach &  Auto-selection \\\hline
\end{tabular}
\caption{Counterparts}\label{tab:methods}
\end{table}
\end{small}

\emph{1}) We first implement the classic fingerprinting-based approach \textsf{CellSense} \cite{IbrahimY12} and a very recent improvement work \textsf{NBL} \cite{MargoliesBBDJUV17}. \textsf{NBL} assumes a prior Gaussian probability of signal strength in divided cell grids. We note that the reasonable size of cell grids in \textsf{NBL} involves the following trade-off: each cell grid should be great enough to contain sufficient MR samples, and yet an excessive size of the grid could alternatively lead to higher localization error (because the center of a greater grid, which is used to approximate the positions of all samples within the grid, leads to a higher error).

\emph{2}) The previous work \textsf{CCR} \cite{ZhuLYZZGDRZ16} implements a pure RaF-based regression model and has demonstrated better localization accuracy than other existing works including the classic work \textsf{CellSense}. Since \textsf{CCR} does not perform knowledge transfer and \textsf{TLoc} performs knowledge transfer on top of RaF-based regressor, we thus name it \textsf{Non-Transfer} in Table \ref{tab:methods}. In addition, we also implement a recent deep neural network-based localization approach, namely \textsf{DeepLoc} \cite{DBLP:conf/gis/ShokryTY18}) as one of the non-transfer learning approaches.

\emph{3}) We are interested in how the adapted STL model is comparable with other transfer learning techniques. Consider that multi-task learning (\textsf{MTL}) is widely used in the transfer learning community \cite{DBLP:journals/ieicet/SimmAS14}, and Supported Vector Regression (\textsf{SVR}) has been used for indoor WiFi localization \cite{DBLP:conf/aaai/ZhengPYP08}. We thus develop two variants of \textsf{TLoc} by using \textsf{MTL} and \textsf{SVR} as the alternative transfer learning techniques. For the three transfer learning-based approaches (STL, \textsf{MTL}, and \textsf{SVR}), we follow \textsf{TLoc} to divide a big area of interest (where each MR data set was sampled) into smaller domains and perform knowledge transfer from source domains to target ones.

We tune the key parameters of the aforementioned counterparts as follows. Firstly,  according to CellSense \cite{IbrahimY12} and DeepLoc \cite{DBLP:conf/gis/ShokryTY18}, we carefully tune the grid size of $50\times 50m^2$ for the best localization precision. In addition, following \cite{DBLP:conf/aaai/ZhengPYP08}, we use the radial basis function (RBF) kernel in \textsf{SVR-Transfer}. Since \textsf{Non-Transfer}, \textsf{MTL}, and \textsf{TLoc} are all RaF-based approaches, we follow the previous works including a benchmark \cite{DBLP:conf/mdm/HuangRZLYZY17} and Non-transferr \cite{ZhuLYZZGDRZ16} to carefully tune the following parameters of RaFs: \emph{1)} the number of trees is set to 200 (to achieve a good trade-off between accuracy and time cost), \emph{2)} the number of used features when looking for the best split is set to $\sqrt{n_f}$ (e.g., $n_f=44$ is the number of total features in 2G GSM MR datasets), and \emph{3)} nodes are expanded until all leaves are pure.

\begin{figure}[th]
	\begin{center}
		\begin{tabular}{c c}
			\begin{minipage}[t]{0.48\linewidth}
				\begin{center}
					\centerline{\includegraphics[height=1.1in]{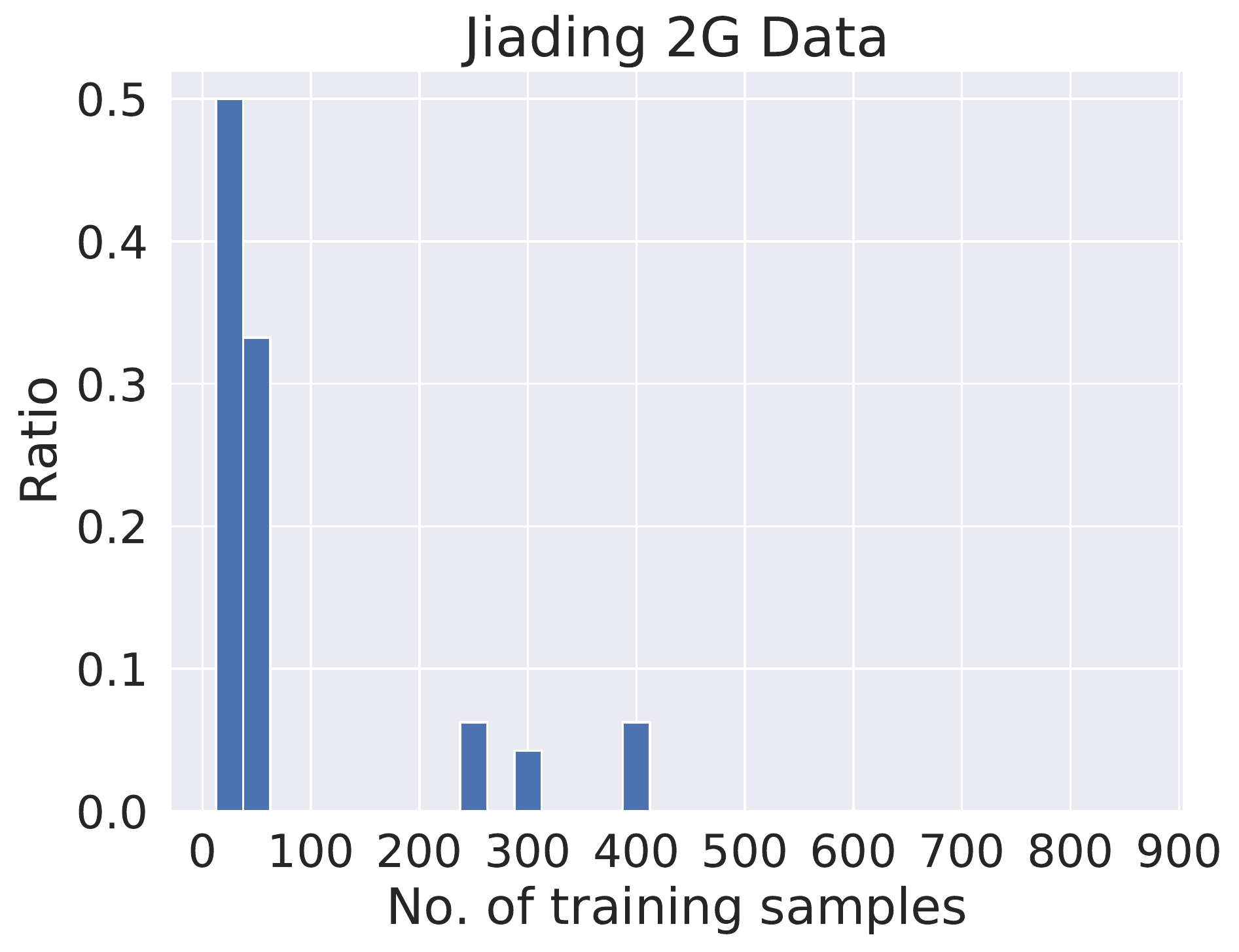}}\vspace{-3ex}
				\end{center}\vspace{-2ex}
			\end{minipage}
			&
			\begin{minipage}[t]{0.48\linewidth}
				\begin{center}
					\centerline{\includegraphics[height=1.1in]{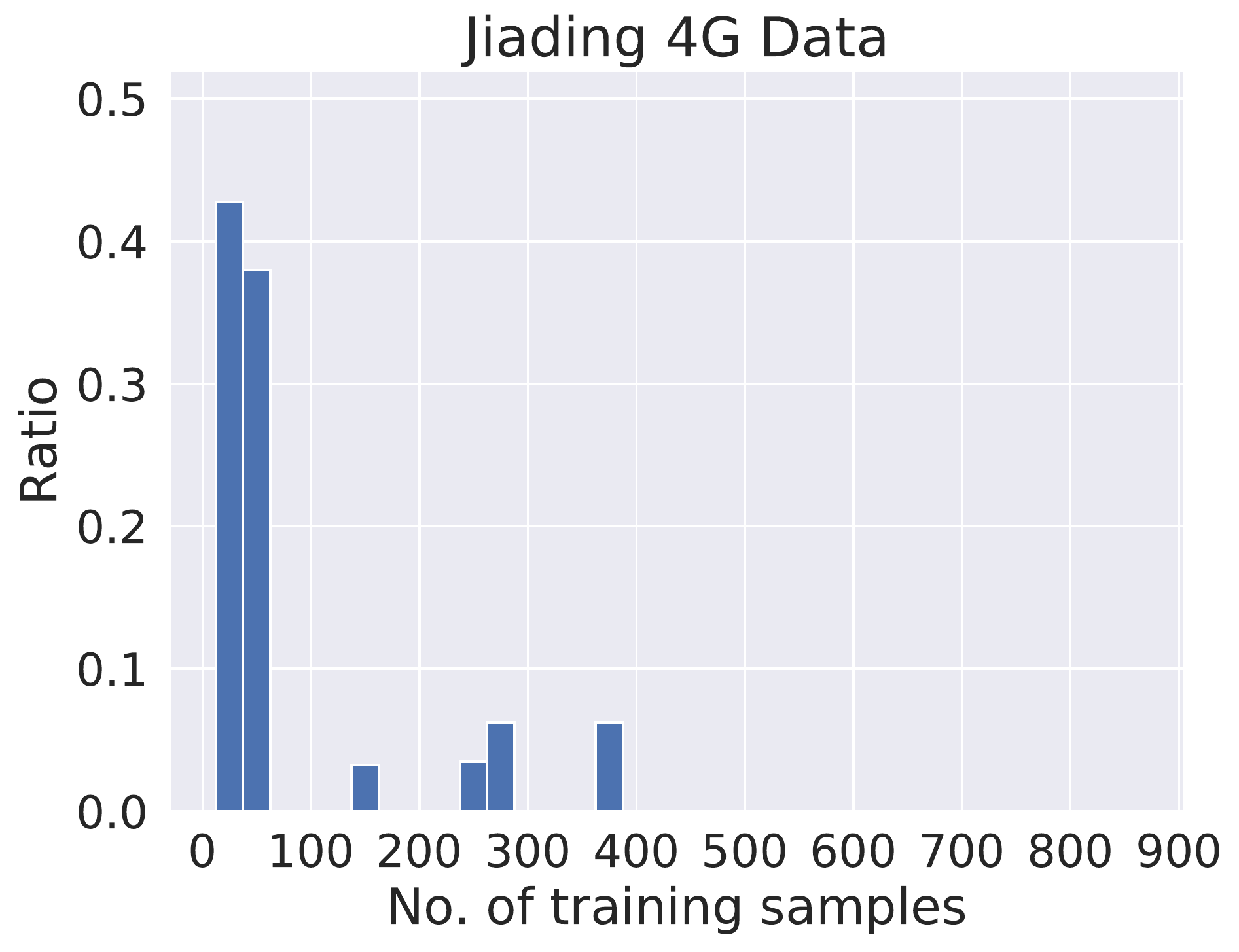}}\vspace{-3ex}
				\end{center}\vspace{-2ex}
			\end{minipage}
		\end{tabular}\vspace{-3ex}
	\end{center}
\caption{Number of Training Samples in those Domains with Low-Accuracy. } \label{fig:transfer_condition}
\end{figure}

Following the work \cite{DBLP:conf/mdm/HuangRZLYZY17}, we adopt the following criteria to empirically determine whether or not a certain domain is treated as a target one. A domain is considered as a target domain, if \emph{1)} the median error of this domain is greater than 30 meters for 4G LTE data or 40 meters for 2G GSM data, and \emph{2)} the number of training samples within this domain is smaller than a threshold, $\tau=50$. From the localization result of non-transfer \textsf{CCR} in each domain (we use \emph{Jiading} datasets for illustration), we find that \emph{1)} the number of training samples in each domain is between the interval from 22 to 864 and \emph{2)} the localization median error is between 8.3 to 86.3 meters. Moreover, we find a strong correlation between the localization error and the quantity of training samples. That is, among those domains with median errors greater than the aforementioned thresholds (i.e., the so-called target domains with low accuracy), 85\% of them contain 50 or even fewer (labelled) MR samples. Figure \ref{fig:transfer_condition} plots the distribution of the number of training samples in the domains of low accuracy. Thus, we empirically set $\tau=50$ for target domains, such that the majority of available domains have improved localization performance by \textsf{TLoc}.


During the evaluation, we adopt 10 times 5-fold cross validation to choose 80\% training and 20\% testing data from each data set \cite{DBLP:conf/ijcai/Kohavi95}, and compare the prediction result of the testing data against ground truth. We compute the prediction error by the Euclidean distance between prediction result and ground truth.

\begin{table}[!hbp]
\small
\scriptsize
\centering
\begin{tabular}{|l|l|}
\hline
\textbf{Parameter} & \textbf{Default Values}\\\hline\hline
Transfer techniques in RaF & Structure transfer \\\hline
Top $k$ source domains & $k=3$ \\\hline
Localization threshold of a target domain (meters) & 40 (2G)/ 30 (4G) \\\hline
$\tau$: Num. of MR samples of a target domain & 50 \\\hline
\% of used MR samples in target/source domains& 80/100 \\\hline
Domain distance weights& $w_{mr}=w_{pos}=0.5$\\\hline
\end{tabular}
\caption{Key Parameters}\label{tab:key_params}
\end{table}

Table \ref{tab:key_params} lists the values of the key parameters in our experiments. We use the default values for the baseline experiments, and vary their values in some appropriate range for sensitivity study. Given the experimental settings above, we mainly evaluate \textsf{TLoc} to study \emph{1)} how \textsf{TLoc} performs against the counterparts (Section \ref{sec:baseline}), \emph{2)} how \textsf{TLoc} is generally beneficial to various transfer learning approaches and other localization schemes (Section \ref{sec:benefit}), \emph{3)} how to meaningfully select source domains (Section \ref{sec:doaminsel}), \emph{4)} how to design an effective measurement of  domain distance (Section \ref{sec:distance}), and finally \emph{5)} how \textsf{TLoc} is sensitive to some key parameters such as the number of source/target MR samples (Section \ref{sec:sens}). After that, we visualize the localization result (Section \ref{sec:vis}) and give the discussion (Section \ref{sec:discuss}).

\subsection{Baseline Study}\label{sec:baseline}
\begin{table*}[tp]
\scriptsize
  \centering
  {  \tiny
    \begin{tabular}{|l|c|c|c|c|c|c|c|c|c|c|c|c|c|c|c|c|c|c|c|c|c|}
    \hline
    \multirow{2}{*}{Dataset}&
    \multicolumn{3}{c|}{Jiading(2G)}&\multicolumn{3}{c|}{Jiading(4G)}&\multicolumn{3}{c|}{Siping(2G)}&\multicolumn{3}{c|}
    {Siping(4G)}&\multicolumn{3}{c|}{Xuhui(2G)}&\multicolumn{3}{c|}{Xuhui(4G)}&\multicolumn{3}{c|}{Urumqi(2G)}\cr\cline{2-22}
    \scriptsize
    &50\%&Mean&90\% &50\%&Mean&90\% &50\%&Mean&90\% &50\%&Mean&90\% &50\%&Mean&90\% &50\%&Mean&90\% &50\%&Mean&90\% \cr
    \hline
    \hline
   NBL \cite{MargoliesBBDJUV17} &53.8&67.4&188.8&51.8&69.3&179.5&42.8&63.0&298.3&43.2&64.9&256.7
   &45.9&59.0& 216.8&32.2&52.4&191.6&58.3&70.2&213.6\cr
   CellCense \cite{IbrahimY12}&55.4&68.7&181.1&55.6&70.6&176.4&44.9&65.7&275.4&45.8&66.3&262.6
   &44.7&60.2&221.3&34.9&55.5&184.3&59.7&71.4&198.7\cr
   DeepLoc \cite{DBLP:conf/gis/ShokryTY18}&37.8&47.3&175.3 &37.2&48.9&184.5&35.5&44.7&219.9 &38.7&49.6&267.5 &31.2&40.3&210.5&27.4&39.8&180.1&44.2&62.9&175.6\cr
   No-Transfer \cite{ZhuLYZZGDRZ16} &38.8&47.6&109.8&35.6&46.5&100.9 &37.5&42.8&119.5 &35.8&41.4&113.7
   &30.0&40.2&113.4 &20.0&34.1&98.3 &45.2&64.3&132.3\cr
   \hline
   MTL \cite{DBLP:journals/ieicet/SimmAS14} &34.3&44.4&80.2&32.1&42.7&79.4&34.3&40.6&89.4&32.2&40.1&77.9
   &28.8&38.9&80.3 &19.5&33.7&96.6 &38.3&60.5&99.7\cr
   SVR-Transfer \cite{DBLP:conf/aaai/ZhengPYP08} &78.4&90.3&79.8&91.8&47.2&167.4&78.4&88.2&145.3&74.5&85.7&159.7
   &59.3&70.3&152.2 &44.8&60.2&149.7 &68.9&81.4&150.7\cr
   TLoc&\textbf{28.1}&\textbf{40.2}&\textbf{72.3}&\textbf{26.3}&\textbf{39.6}&\textbf{69.8}&\textbf{28.8}&\textbf{39.7}&\textbf{69.2}&\textbf{23.2}&\textbf{37.4}&\textbf{67.4}&\textbf{27.7}&\textbf{37.5}&\textbf{72.5}&\textbf{18.9}&\textbf{32.4}&\textbf{69.5}&\textbf{35.4}&\textbf{49.1}&\textbf{92.8}\cr
   \hline
   \end{tabular}}
  \caption{Baseline Experiment}\vspace{-2ex}
  \label{tab:performance_comparison}\vspace{-2ex}
\end{table*}
\normalsize

We first report the position recovery errors of seven Telco recovery approaches. In Table \ref{tab:performance_comparison}. We show the median, mean and 90\% errors (denoted by 50\%, $M_e$, and 90\%) in cases of using 2G and 4G network data, respectively. From Table \ref{tab:performance_comparison}, we have the following findings.

First, \textsf{TLoc} achieves the least errors among the seven approaches on all data sets. For example, in \emph{Siping} 2G GSM dataset, the median error of \textsf{NBL}, \textsf{CellSense}, \textsf{DeepLoc}, \textsf{Non-Transfer}, \textsf{MTL}, \textsf{SVR-Transfer}, and \textsf{TLoc} algorithms is 42.8, 44.9, 35.5, 37.5, 34.3, 78.4 and 28.8 meters, respectively. Such result indicates that \textsf{TLoc} outperforms the \textsf{Non-Transfer} approach by 23.2\%. Similar situation occurs on other data sets. Among the seven algorithms, the three RaF-based algorithms, including \textsf{TLoc}, \textsf{MTL}, and \textsf{Non-Transfer}, lead to better accuracy than \textsf{SVR}-based and fingerprint-based algorithms. Moreover, \textsf{Non-Transfer}, i.e., the RaF-based localization approach, indicates the comparable localization accuracy to \textsf{DeepLoc}. 

Second, the 4G LTE data sets exhibit lower errors than the 2G GSM data sets. For example, in \emph{Siping} 4G LTE data set, \textsf{TLoc} has the median error of 23.20 meters, 16.88\% lower localization error when compared to \emph{Siping} 2G GSM dataset. By carefully checking the database of base stations, we find that the 4G LTE base stations are deployed more densely than the 2G GSM stations. In addition, \emph{Siping} campus is located at the urban areas in Shanghai with denser deployment of base stations than {\emph{Jiading}} campus in sub-urban areas in Shanghai. Thus, the localization errors on \emph{Siping} data sets, including 2G GSM and 4G LTE, are smaller than those on \emph{Jiading} data sets.

Third, in terms of the localization performance of \textsf{TLoc} against the two fingerprint-based methods \textsf{CellSense} and \textsf{NBL}, \textsf{TLoc} consistently outperforms the two fingerprint-based methods on all data sets. In addition, \textsf{NBL} and \textsf{CellSense} exhibit very similar curves on all data sets, though \textsf{NBL} leads to slightly lower errors than \textsf{CellSense}. This result is consistent with the one reported by the evaluation of \textsf{NBL}. Note that due to the similar curves between \textsf{CellSense} and \textsf{NBL}, in the rest of this section, we mainly choose \textsf{NBL} as the representative implementation of a fingerprint-based method.

Fourth, although \textsf{TLoc} is used to overcome the data scarcity issue, In Table \ref{tab:performance_comparison}, it is interesting to see how \textsf{TLoc} generally performs in diverse domains, e.g., those with sufficient MR samples (e.g., \emph{Xuhui} dataset) and those located in a city such as \emph{Urumqi}, which has a rather different distribution of base stations from the large urban city \emph{Shanghai}. From the results of \emph{Xuhui} and \emph{Urumqi} data sets, we have two findings. First for the domains in \emph{Xuhui}, \textsf{TLoc} again consistently outperforms \textsf{Non-Transfer}, although relatively small improvement when compared with the results in \emph{Jiading} and \emph{Siping} data sets. Second, for the domains in \emph{Urumqi}, it is not surprising that the localization error for the \emph{Urumqi} data set is much higher than that for the \emph{Xuhui} data set, mainly due to the rather sparse deployment of base stations in \emph{Urumqi}. Nevertheless, in the \emph{Urumqi} data set, \textsf{TLoc} still leads to a significant reduction of localization errors over \textsf{Non-Transfer}.

Finally, in terms of the accuracy of the RaF-based transfer learning approaches, we find that \textsf{TLoc} outperforms \textsf{MTL} in all data sets. It is mainly because \textsf{MTL} learns the tasks for both source and target domains, and yet \textsf{TLoc} adaptively tunes the split thresholds on RaF nodes by the MR samples in target domains. In addition, those three RaF-based algorithms, including \textsf{TLoc}, \textsf{MTL}, and even \textsf{Non-Transfer}, all achieve much better accuracy than \textsf{SVR}-based and fingerprint-based algorithms, consistent with the benchmark \cite{DBLP:conf/mdm/HuangRZLYZY17}. The main reason is that it is hard for \textsf{SVR} to select an appropriate kernel function for the nonlinear feature space of MR samples. Meanwhile the hierarchical tree in RaF works very well to model the spatial structure: from a big area \cite{ZhuLYZZGDRZ16} to divided small domains.


\subsection{Benefits of TLoc}\label{sec:benefit}
\begin{figure}[th]
	\hspace{-3ex}
	\begin{center}
		\begin{tabular}{c c}
			\begin{minipage}[t]{0.47\linewidth}
				\begin{center}
					\centerline{\includegraphics[width=1.5in]{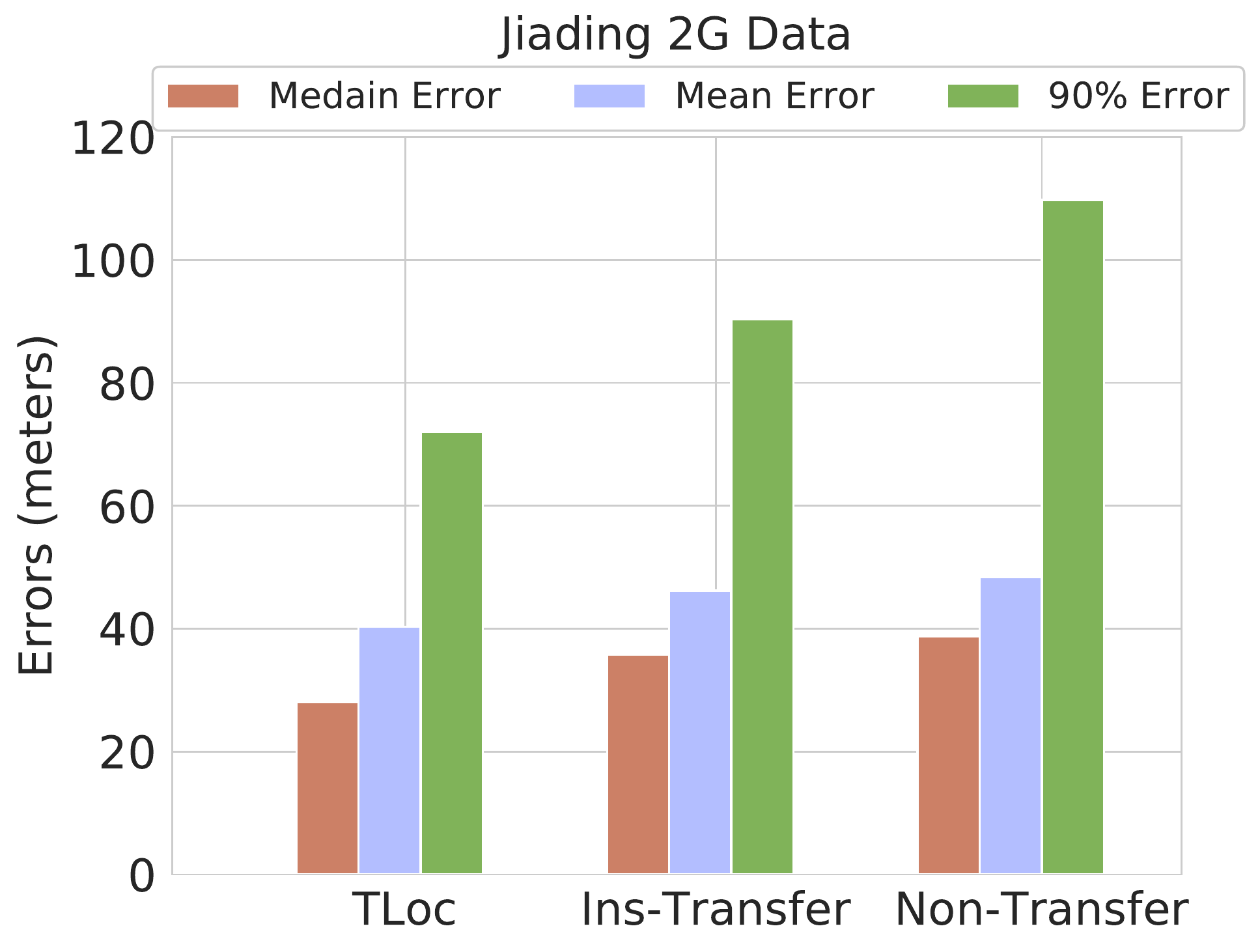}}
				\end{center}
			\end{minipage}
			&
			\begin{minipage}[t]{0.47\linewidth}
                    \begin{center}
					\centerline{\includegraphics[width=1.5in]{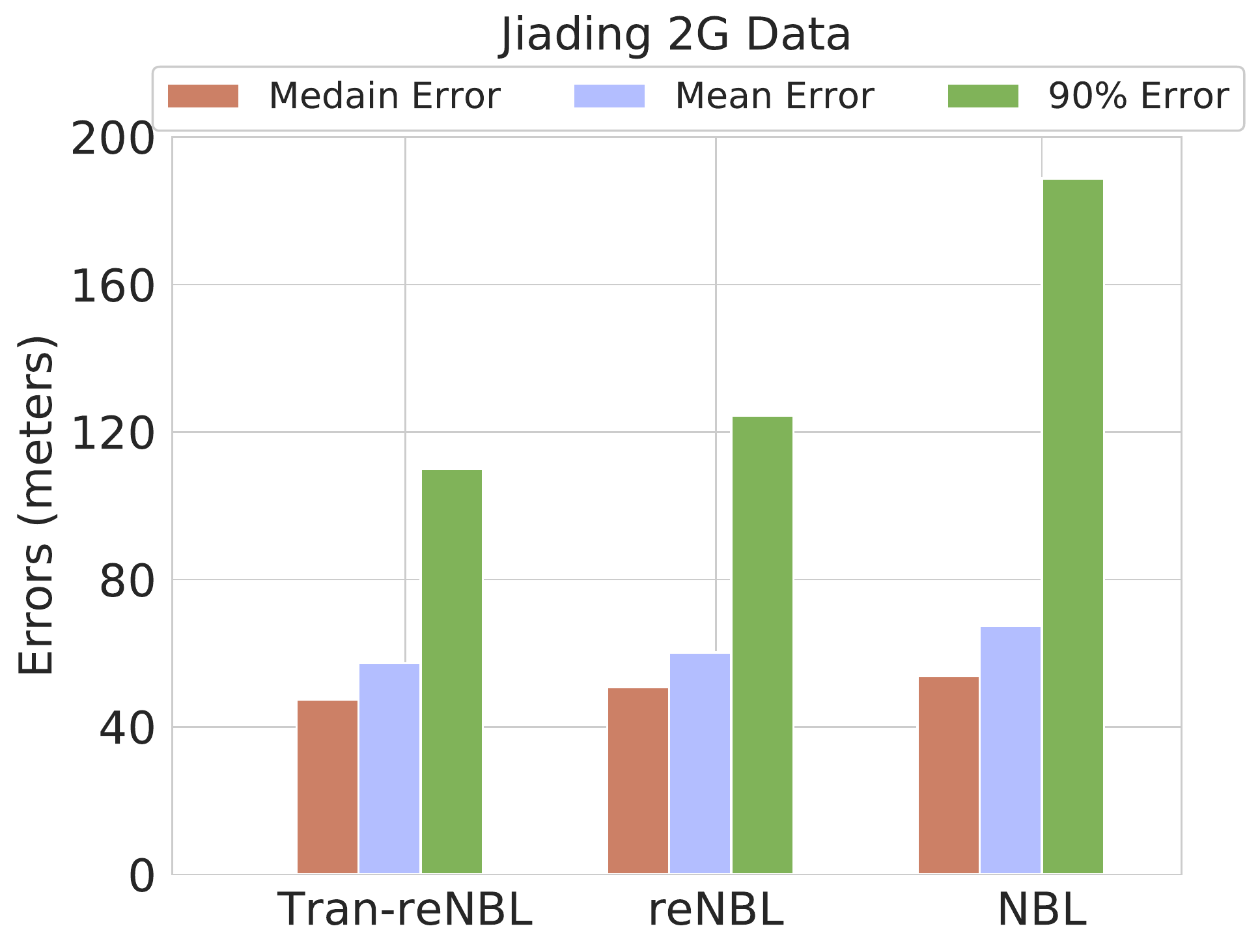}}
				\end{center}
			\end{minipage}
		\end{tabular}\vspace{-3ex}
		\caption{Benefits of TLoc (from left to right). (a) Instance-based Transfer, (b) Fingerprinting-based localization.}
		\label{exp:benefit}\vspace{-3ex}
	\end{center}
\end{figure}

\textbf{Benefit to Instance-based Transfer Learning:} Beyond the model-based STL used by \textsf{TLoc}, we believe that the top-$k$ source domains can offer benefits to other transfer learning techniques, e.g., instance-based transfer. To this end, based on the selected source domains, we mix the MR samples from both source and target domains to train a RaF regression model for the target domains. This approach can be intuitively treated as instance-based transfer, namely \textsf{Ins-Transfer}. Figure \ref{exp:benefit}(a) plots the results of \textsf{Non-Transfer}, \textsf{Ins-Transfer} and \textsf{TLoc}. Both \textsf{Ins-Transfer} and \textsf{TLoc} lead to lower errors than \textsf{Non-Transfer}. These results verify the benefits of using the top-$k$ similar source domains.

\textbf{Benefit to Fingerprinting-based Localization:} In this experiment, we explore the potential of applying the techniques developed for \textsf{TLoc} to fingerprinting-based methods, e.g., \textsf{NBL} \cite{MargoliesBBDJUV17}. Similar to \textsf{TLoc}, we divide the area of interest into multiple domains and perform the representation of MR features and position labels as before. Next, for the MR features and positions within each domain, we follow \textsf{NBL} to perform the fingerprinting-based position recovery. We name the \textsf{NBL} method in relative coordinate space as \textsf{reNBL}. Based on the \textsf{reNBL}, we implement the instance-based transfer, namely \textsf{Tran-reNBL}, by first mixing the training samples from source and target domains and then performing position recovery by \textsf{reNBL}. We compare \textsf{NBL} and the two variants \textsf{reNBL} and \textsf{Tran-reNBL} in Figure \ref{exp:benefit}(b). As shown in this figure, the instance transfer in \textsf{Tran-reNBL} does lead to the lowest localization error among the three methods as expected.

\begin{figure}[th]
	\hspace{-8ex}
	\begin{center}
		\begin{tabular}{c c}
			\begin{minipage}[t]{0.47\linewidth}
				\begin{center}
					\centerline{\includegraphics[height=1.2in]{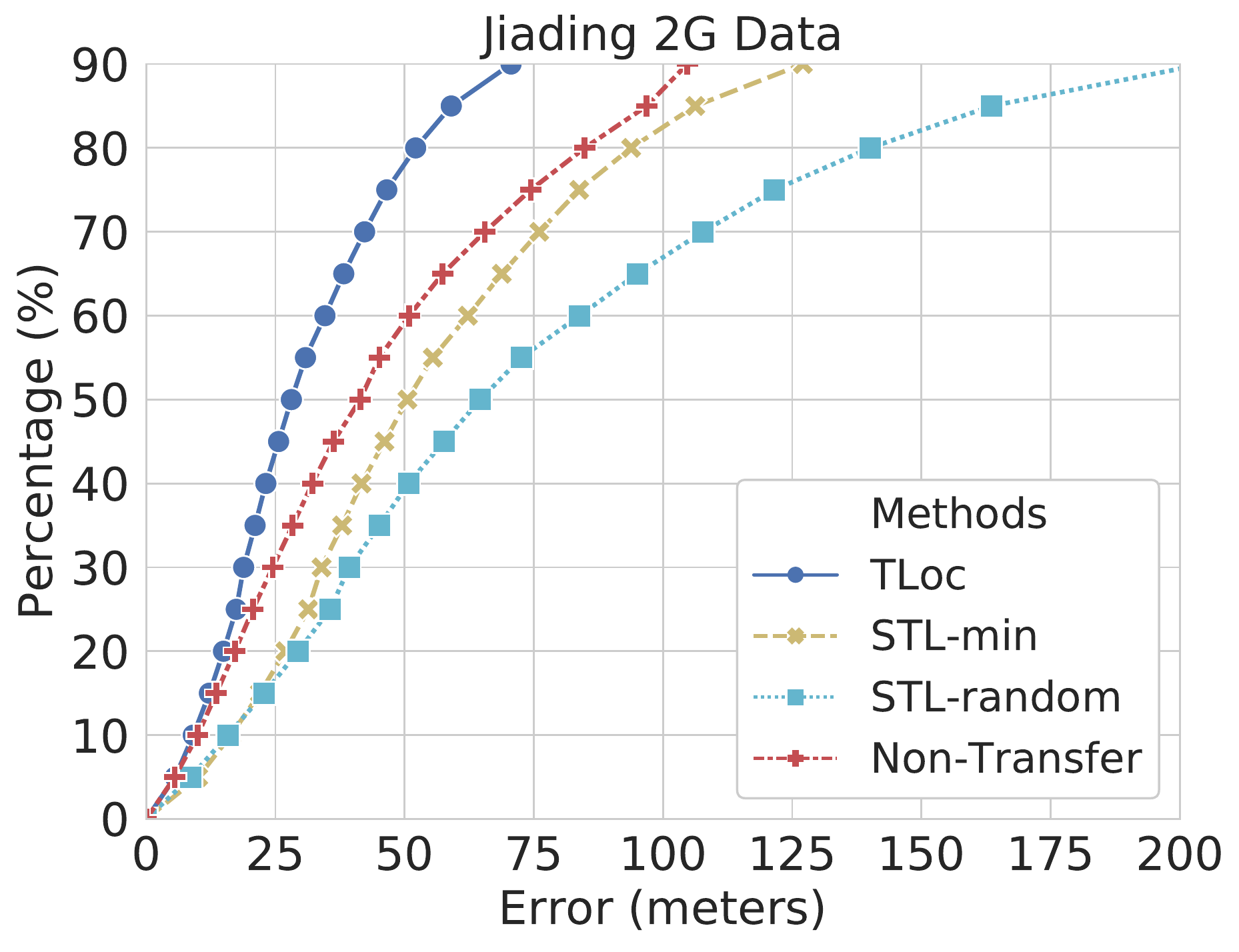}}\vspace{-5ex}
				\end{center}
			\end{minipage}
			&
			\begin{minipage}[t]{0.47\linewidth}
				\begin{center}
					\centerline{\includegraphics[height=1.2in]{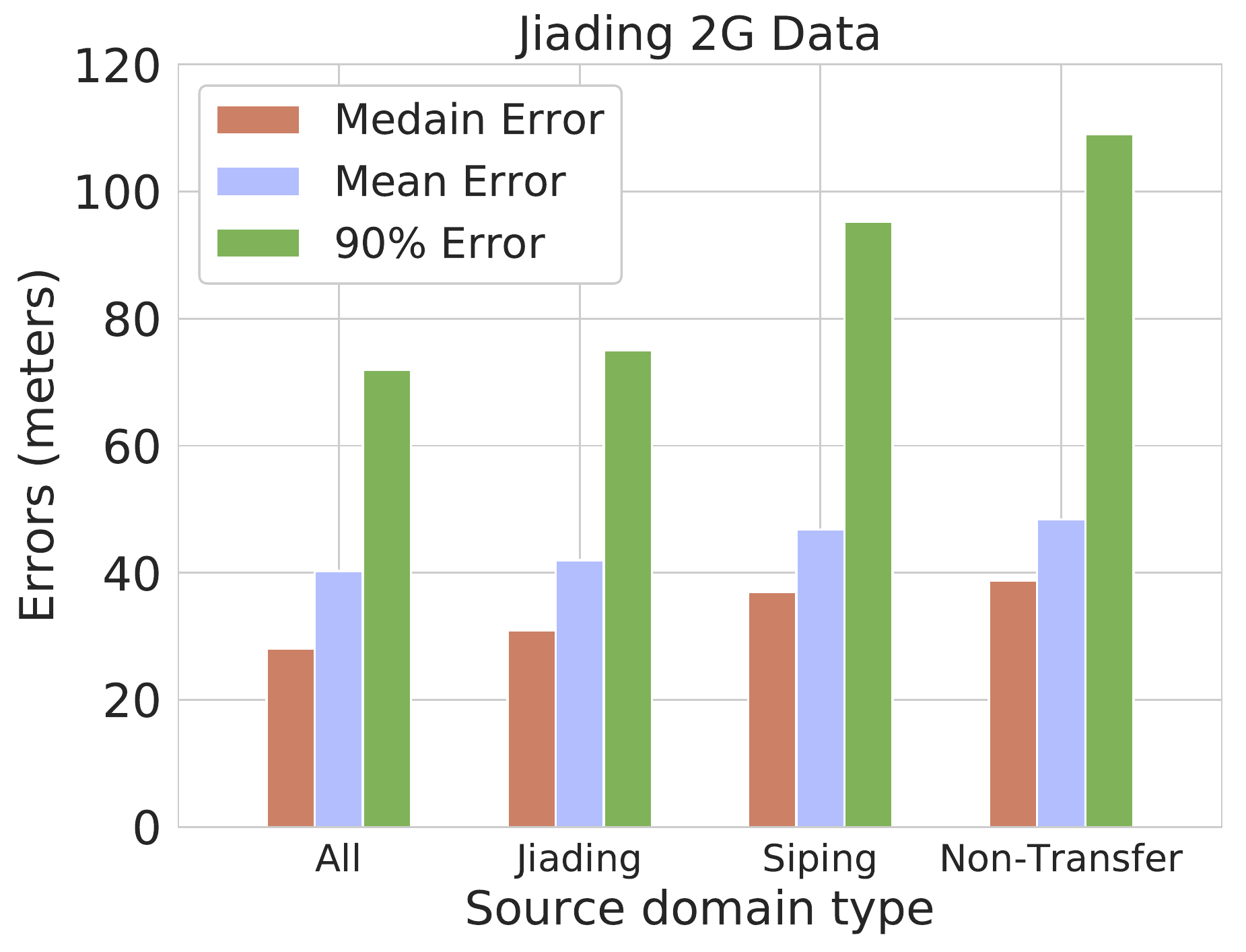}}\vspace{-5ex}
				\end{center} 
			\end{minipage}
			\end{tabular}\caption{Source Domain Selection (from left to right). (a) Four approaches, (b) Areas of source domains}\vspace{-3ex}\label{exp:srcdomain}
	\end{center}
\end{figure}

\subsection{Source Domain Selection}\label{sec:doaminsel}
\textbf{Domain Selection}: First we compare the proposed approach of selecting the top-$k$ most similar source domains against two alternative approaches: \emph{1)} \textsf{STL\_min} selects the top-$k$ domains with the least prediction error (achieved by \textsf{Non-Transfer}), and \emph{2)} \textsf{STL\_random} randomly selects $k$ source domains. After these source domains are selected, we adopt STL to transfer knowledge from source domains to target ones. As shown in Figure \ref{exp:srcdomain}(a), both \textsf{STL\_min} and \textsf{STL\_random} even lead to higher errors than \textsf{Non-Transfer}. The result verifies the necessity of carefully selecting the most similar source domains. Otherwise, those dissimilar source domains, e.g., those selected by \textsf{STL\_min} and \textsf{STL\_random} even harm the localization accuracy of target domains.

\textbf{Domain Distance}: Motivated by the result above, we are further interested in the effect of selected source domains by various domain distance. In Table \ref{tab:negative_transfer}, the target domains of \emph{Jiading} 2G data set are divided into 5 groups according to the average domain distance of Top-$k$ (= 3) source domains. For each group, we compute the average median errors on target domains before transfer and after transfer. From this table, we have the following findings. \emph{1)} A source domain with lower distance (a.k.a higher similarity) to a target domain leads to a more positive transfer effect with lower localization errors. It means that using similar source domains does improve the localization accuracy of target domains. \emph{2)} When the domain distance is greater than 0.95 (though the proportion of such target domains is trivially 1.7\%), it indicates the selected source domains are rather dissimilar to the target domain. Such source domains result in a negative transfer effect and higher localization errors, consistent with the result in Figure \ref{exp:srcdomain}(a). Thus, we can empirically set a pre-defined threshold of domain distance, e.g., 0.95, to prune such dissimilar source domains. In this way, we can ensure that the selected source domains are truly similar to target ones and thus avoid the negative transfer effect of dissimilar source domains. 

\begin{table}[!hbp]
\scriptsize
\begin{tabular}{|l|c|c|c|c|c|}
\hline
\multirow{2}{*}{\textbf{\begin{tabular}[c]{@{}l@{}}Median Error on  \\ Target Domain (meters)\end{tabular}}} & \multicolumn{5}{c|}{\textbf{Avg. Domain Distance of Source Domains}} \\ \cline{2-6}
 & $<$0.4 & 0.4-0.6 & 0.6-0.8 & 0.8-0.95 & $>$0.95 \\ \hline
Before Transfer & 49.3 & 51.3 & 48.9 & 51.9 & 50.4 \\ \hline
After Transfer & 34.1 & 35.7 & 37.7 & 46.2 & 51.2 \\ \hline \hline
\% of Target Domains & 15.3 & 25.6 & 42.8 & 14.6 & 1.7 \\ \hline
\end{tabular}\caption{Source Domain Effects of Varying Domain Distances between Source and Target Domains: \emph{Jiading} 2G GSM Data Set.}\label{tab:negative_transfer}
\end{table}

\textbf{Areas of Source Domains}: Third, we are interested in the areas where selected source domains are located. To this end, we purposely select source domains from 1) all three areas in Shanghai (all), 2) {\emph{Jiading}} alone, and 3) \emph{Siping} alone. In Figure \ref{exp:srcdomain} (b), the source domains from all areas lead to the least errors, and \textsf{Non-Transfer} suffers from the highest errors. Specifically, for the target domains in \emph{Jiading} 2G data set, if we select source domains from all three areas in Shanghai, we can find that 11.1\% selected source domains are from \emph{Xuhui}, 28.4 \% source domains are from \emph{Siping}, and 60.5\% source domains are from \emph{Jiading}. These numbers indicate that most of source domains and the corresponding target domains are within the same area, but still a small number of source domains are from the two other areas. If we only select the source domains from the same area where the target ones are located, those source domains from other areas could be missed. In addition, as shown in Figure \ref{exp:srcdomain} (b), the source and target domains within the same area \emph{Jiading} can achieve less errors than those across areas, i.e., the target domains in \emph{Jiading} but the source domains in \emph{Siping}. It is because among those similar source domains for a certain target domain, most of them are within the same area, and a small number of them are from other areas, consistent with Table \ref{tab:negative_transfer}.

\begin{figure}[th]
	\hspace{-3ex}
	\begin{center}
		\begin{tabular}{c c}
			\begin{minipage}[t]{0.47\linewidth}
				\begin{center}
\centerline{\includegraphics[height=3.2cm]{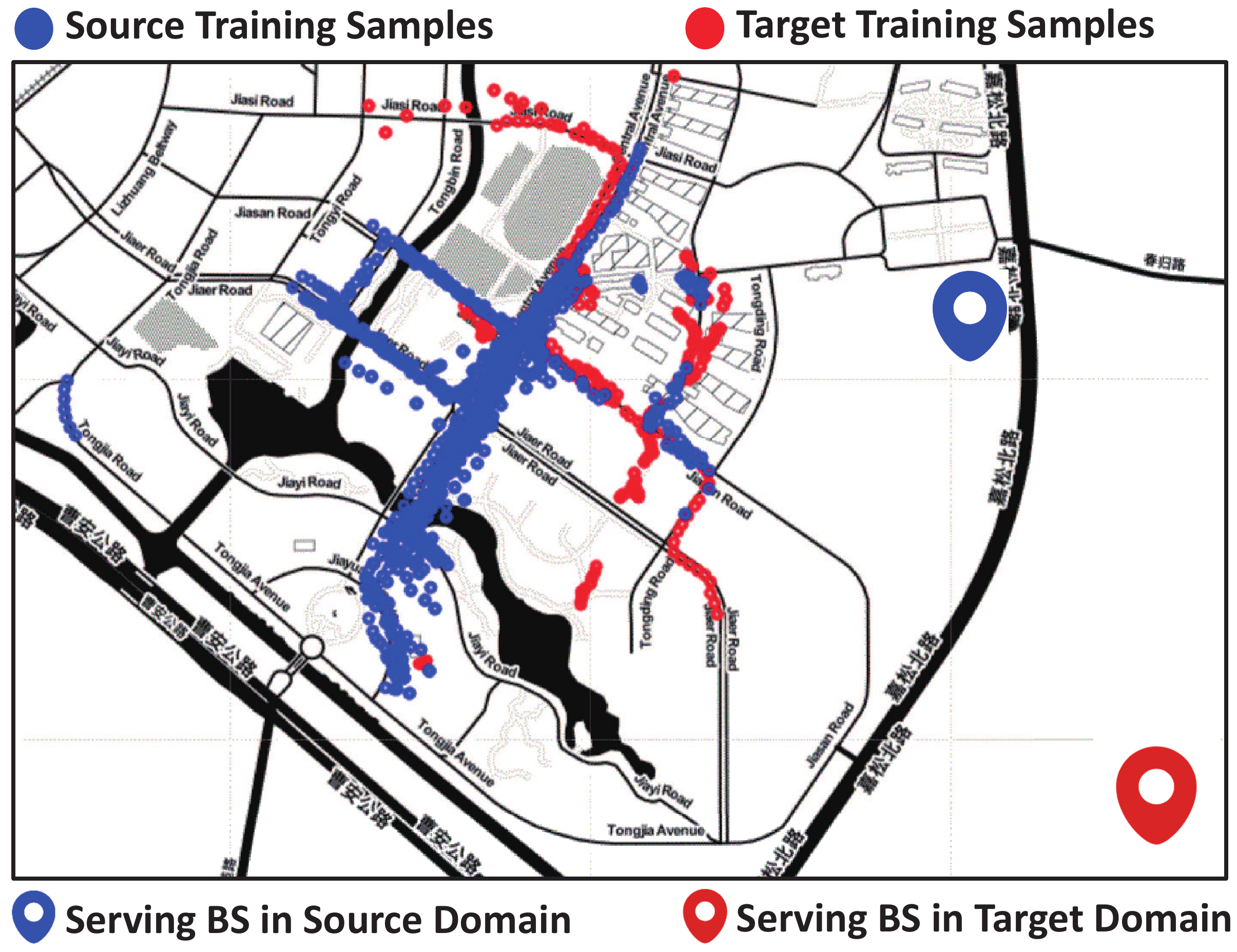}}
				\end{center}
			\end{minipage}
			&
			\begin{minipage}[t]{0.47\linewidth}
                    \begin{center}
					\centerline{\includegraphics[height=3.2cm]{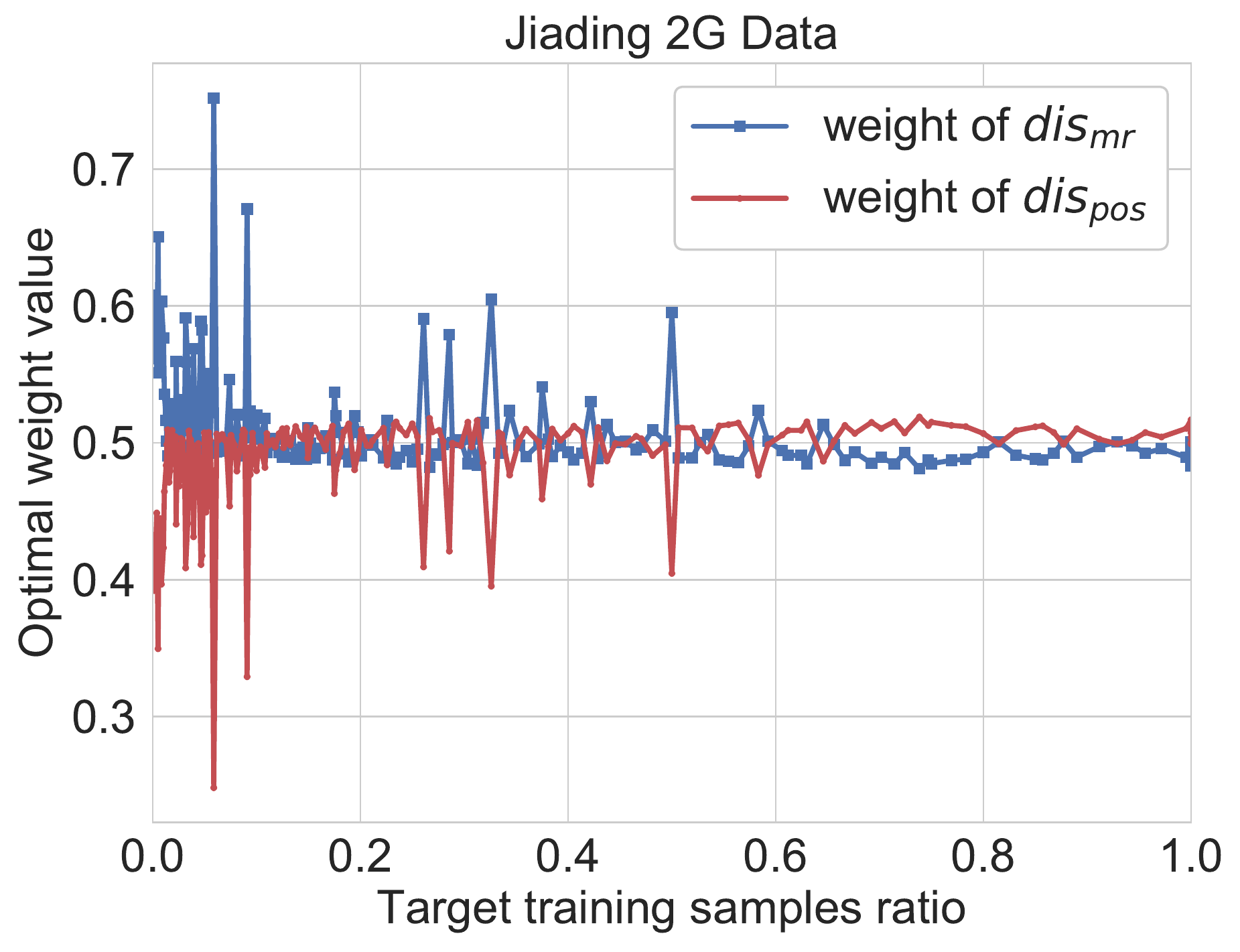}}
				\end{center}
			\end{minipage}
		\end{tabular}\vspace{-3ex}
		\caption{From left to right: (a) Domain Intersection, (b) Weight Tuning}
		\label{exp:intersection}\vspace{-3ex}
	\end{center}
\end{figure}

\textbf{Source-Target Domains within the Same Area}: Differing from the experiment in Figure \ref{exp:srcdomain} (b) above, we now evaluate \textsf{TLoc} on the source-target domains within partially overlapping areas. Figure \ref{exp:intersection} illustrates an example scenario for two specially chosen domains in our \emph{Jiading} 2G dataset: the MR samples (blue dots) in a certain source domain and those (red dots) in a target domain are partially co-located within the same road segments. Given this scenario, we purposely study various approaches to select MR samples from the source domain, and evaluate the performance of \textsf{TLoc}. From Table \ref{tab:intersection}, we find that simply selecting the source samples only from the overlapping road segments incurs the highest errors. It is mainly because the source and target samples even within the same road segments could exhibit very different signal features and relative position coordinates (because MR samples within the same road segments could be connected to various serving base stations). Instead, via the STL scheme, \textsf{TLoc} adapts the RaF regression model built from the source domain to the target one, leading to the least error. This experiment clearly indicates the advantages of \textsf{TLoc} over the approach that simply selects those source domains located at the same road segments as the target ones.

\begin{table}[!hbp]
\scriptsize
\centering
\begin{tabular}{|c|c|c|c|}
\hline
\textbf{Types of Source Samples} & \textbf{Median}& \textbf{Mean}& \textbf{90\%}\\\hline
All samples in source & 40.6&51.4&108.7\\\hline
Samples in intersection area& 42.4&52.5&110.4\\\hline
Samples in non-intersection area& 41.6&52.9&112.3\\\hline
\hline
\textbf{Non-Transfer} & 42.5&53.7&105.2\\\hline
\textbf{TLoc with Source Selection}& 33.8&45.3&94.4\\\hline
\end{tabular}
\caption{Effect of Intersection between Two Different Domains.}\label{tab:intersection}
\end{table}

\textbf{Trade-off between Localization Errors and Time Cost}: First, by varying the number $k$, we study the effect of the number $k$ on the median error and running time of \textsf{TLoc} (due to space limit, this figure is not shown). The experimental result indicates that a greater number $k$ in general leads to decreased errors, but the curve remains rather stable for $k>3$. The errors even become slightly higher when $k$ reaches 5. It is mainly because a greater number $k$ indicates less similarity between source and target domains. A dissimilar source domain may lead to negative transfer effect. In terms of the time efficiency of \textsf{TLoc} measured by the training and prediction time, as the number $k$ grows, more training samples are used by the model, leading to more running time. Thus, to balance the trade-off between time efficiency and model accuracy, we by default set $k=3$.

Next, consider that \textsf{TLoc} requires pairwise domain distance, incurring non-trivial computing overhead. To overcome this issue, we apply the technique of Locality Sensitive Hash (LSH) \cite{DBLP:conf/stoc/IndykM98} to efficiently approximate the domain distance. As shown in Table \ref{tab:lsh}, though LSH is only an approximation approach, it can still achieve acceptable localization errors (e.g., 11.1\% higher median error) while the time cost is greatly reduced by 4.58$\times$.

\begin{table}[th]
\scriptsize
\centering
\begin{tabular}{|c|c|c|c|c|}
\hline
\multicolumn{1}{|c|}{\multirow{2}{*}{\textbf{\begin{tabular}[c]{@{}c@{}}Source Selection \\ Criterion\end{tabular}}}} & \multicolumn{3}{c|}{\textbf{Localization Error (meters)}}         & \multicolumn{1}{c|}{\multirow{2}{*}{\textbf{\begin{tabular}[c]{@{}c@{}}Avg. time per\\Target domain (ms)\end{tabular}}}} \\ \cline{2-4}
\multicolumn{1}{|c|}{}                                                                                                & \textbf{Median} & \textbf{Mean} & \textbf{90\% } & \multicolumn{1}{c|}{}                                                                                                      \\ \hline
Domain Distance   & 28.1             & 40.3          & 72.3         & 1657  \\ \hline
LSH Approximation & 32.4             & 47.7          & 90.6         & 362    \\ \hline
\end{tabular}
\caption{Trade-off between Localization Errors and Time Cost.}\label{tab:lsh}\vspace{-4ex}
\end{table}

\subsection{Domain Distance}\label{sec:distance}

\textbf{Ablation Study of Domain Distance:} Recall that the domain distance is computed by integrating MR feature distance $dis_{mr}$ and relative position distance $dis_{pos}$, and the MR feature distance $dis_{mr}$ is further computed by the weighted items $dis_{mr}^{rssi}$ and $dis_{mr}^{sig}$. Thus, to study the effect of each item, Table \ref{tab:params} first uses $dis_{mr}^{rssi}$, $dis_{mr}^{sig}$, $dis_{mr}$, $dis_{pos}$ alone, and then various combinations of these items to compute domain distance for source domain selection. First, using $dis_{mr}^{rssi}$ alone leads to lower errors than using $dis_{mr}^{sig}$ alone, indicating that $dis_{mr}^{rssi}$ makes a major contribution to $dis_{mr}$. Second, $dis_{mr}$ leads to slightly lower errors than $dis_{pos}$. Finally, it is not surprising that source selection by the distance integrating the weighted $dis_{mr}$ and $dis_{pos}$ leads to the best result.

\begin{table}[!hbp]
\scriptsize
\centering
\begin{tabular}{|c|c|c|c|}
\hline
\textbf{Domain Distance} & \textbf{Medain}     & \textbf{Mean}    & \textbf{90\%}    \\ \hline
$dis_{mr}^{rssi}$ & 33.6&50.9&82.7\\\hline
$dis_{mr}^{sig}$ & 39.4&58.2&99.3\\\hline
$dis_{mr}$ & 32.5&46.4&78.7\\\hline
$dis_{pos}$ & 34.3&49.2&82.5\\\hline
\hline
$0.5*dis_{mr}+0.5*dis_{pos}$ & 28.1&40.2&72.3\\\hline
$0.67*dis_{mr}+0.33*dis_{pos}$  & 31.5&44.4&76.2\\\hline
$0.33*dis_{mr}+0.67*dis_{pos}$ & 33.4&48.6&79.6\\\hline
\end{tabular}
\caption{Ablation Study of Domain Distance: {\emph{Jiading}} 2G GSM Data Set.}\label{tab:params}
\end{table}

In terms of the weights $w_{mr}$ and $w_{pos}$ (See Equation \ref{eq:overalldist} in Section \ref{sec:source}), we study the effect of weight setting on the errors of \textsf{TLoc}. As shown in Table \ref{tab:params}, using either $dis_{mr}$ or $dis_{pos}$ alone, i.e., $w_{mr} = 1.0$ or $w_{pos} = 1.0$, cannot lead to the least error. Instead, the equal weights  $w_{mr} = w_{pos} = 0.5$ lead to the best result. It makes sense because the position recovery model maps MR features to associated positions. Thus, in general, $dis_{mr}$ and $dis_{pos}$ leads to roughly equal importance for domain distance $dist(D,D')$.

We note that the weight setting should be adaptive to the ratio of MR samples between target domains and source ones. To this end, for a given ratio of MR samples between a target domain and source domains, we empirically tune the weights $w_{mr}$ and $w_{pos}$ which lead to the least prediction error, and plot the weight against the MR sample ratio in Figure \ref{exp:intersection}(b). When the ratio is close to 0.0 (indicating that the target domain has very few labelled MR samples), $w_{mr}$ values are typically greater than $w_{pos}$. It is because the domain distance mainly depends upon MR features instead of MR positions (due to the ratio equal to 0.0, i.e., very few MR position labels in target domains). As the ratio becomes greater, i.e., more target labelled samples, $w_{mr}$ remains stabilize equal around 0.5, consistent with Table \ref{tab:params}.

\textbf{Number of Trajectories:} Recall that relative position distance is dependent on the number of trajectories in domains. Thus, to study the effect of the number of trajectories on localization errors, in Table \ref{tab:traj_num}, we divide all target domains into three groups according to the number of trajectories. For each target domain in a group, we compute the distance between this target domain and a certain source domain, then use the distance as the criterion for source domain selection, and finally compute the average median error for all target domains in this group. From this table, the group with more trajectories corresponds to lower localization errors. It is mainly because in our datasets, the group with more trajectories indicates a higher spatial coverage rate of MR samples in target domains. Moreover, more trajectories in target domains indicate more significant contribution of the weight $w_{pos}$ and lead to low errors, which is consistent with the result in Figure \ref{exp:sense}(a).

\begin{table}[!hbp]
\scriptsize
\centering
\begin{tabular}{|c|c|c|c|}

\hline
\multirow{2}{*}{\textbf{\begin{tabular}[c]{@{}c@{}}No. of Traj \\ in Target\end{tabular}}}& \multicolumn{3}{c|}{\textbf{Median Error on Target (meters)}} \\ \cline{2-4}
 & \textbf{\begin{tabular}[c]{@{}c@{}}Source Selection by \\$dist(D,D')$ \end{tabular}} & \textbf{\begin{tabular}[c]{@{}c@{}}Source Selection by \\ $dis_{pos}$ \end{tabular}} & \textbf{Non-Transfer} \\ \hline
1-4 & 42.2 & 50.1 & 62.6 \\ \hline
5-8 & 34.2 & 39.3 & 55.3 \\ \hline
8+ & 27.8 & 32.7 & 45.2 \\ \hline
\end{tabular}
\caption{Effect of Number of Trajectories in Domain Distance: {\emph{Jiading}} 2G GSM Data Set.}\label{tab:traj_num}
\end{table}\vspace{-2ex}

\subsection{Sensitivity Study}\label{sec:sens}
In this section, we vary the values of several key parameters and study the performance of \textsf{TLoc}.

\textbf{Transfer Learning Techniques in Random Forests:} Recall that we adapt the Structure Transfer (STL) technique for model transfer. Besides STL, the previous work \cite{DBLP:journals/pami/SegevHMCE17} proposed two other model transfer algorithms: Structure Expansion/Reduction (SER) and MIX. Here, SER searches greedily for locally optimal modifications of each tree structure by trying to locally expand or reduce the tree around individual nodes, and MIX utilizes a majority vote on the decision trees transferred by either STL or SER. As shown in Table \ref{tab:tl-tech}, we evaluate the effectiveness of these three model transfer techniques. STL leads to the lowest localization errors and SER suffers from the highest errors. It is mainly because the selected source domains are with the highest similarity with the target domain, and STL does not significantly update the DTs trained from source domains. These results are consistent with the previous work \cite{DBLP:journals/pami/SegevHMCE17}, where
source and target images do share the similar geometric shapes though with various inverted colors and other features. In addition, the running time of STL is much faster than STL and MIX due to the trivial update of the node thresholds in the decision trees of STL. Therefore, we implement our model transfer by STL.

\begin{table}[th]
\scriptsize
\centering
\begin{tabular}{|c|c|c|c|c|}
\hline
\multicolumn{1}{|c|}{\multirow{2}{*}{\textbf{\begin{tabular}[c]{@{}c@{}}Transfer Learning \\ Techniques \end{tabular}}}} & \multicolumn{3}{c|}{\textbf{Localization Error (meters)}}         & \multicolumn{1}{c|}{\multirow{2}{*}{\textbf{\begin{tabular}[c]{@{}c@{}}Avg. Training Time \\per Target domain (s)\end{tabular}}}} \\ \cline{2-4}
\multicolumn{1}{|c|}{}                                                                                                & \textbf{Median} & \textbf{Mean} & \textbf{90\% } & \multicolumn{1}{c|}{}                                                                                                      \\ \hline
STL  & 28.1             & 40.2         & 72.3         & 14.2  \\ \hline
SER & 33.4           & 48.7          & 80.4         & 25.9    \\ \hline
MIX & 31.1             & 43.3         & 77.9         & 53.4  \\ \hline
\end{tabular}
\caption{Effect of Different Transfer Learning Techniques in Random Forest: {\emph{Jiading}} 2G GSM Data Set. (STL: Structure Transfer, SER: Structure Expansion/Reduction)}\label{tab:tl-tech}\vspace{-4ex}
\end{table}

\begin{figure*}[th]
	\hspace{-8ex}
	\begin{center}
		\begin{tabular}{c c c c}
            \begin{minipage}[t]{0.25\linewidth}
				\begin{center}
					\centerline{\includegraphics[height=1.16in]{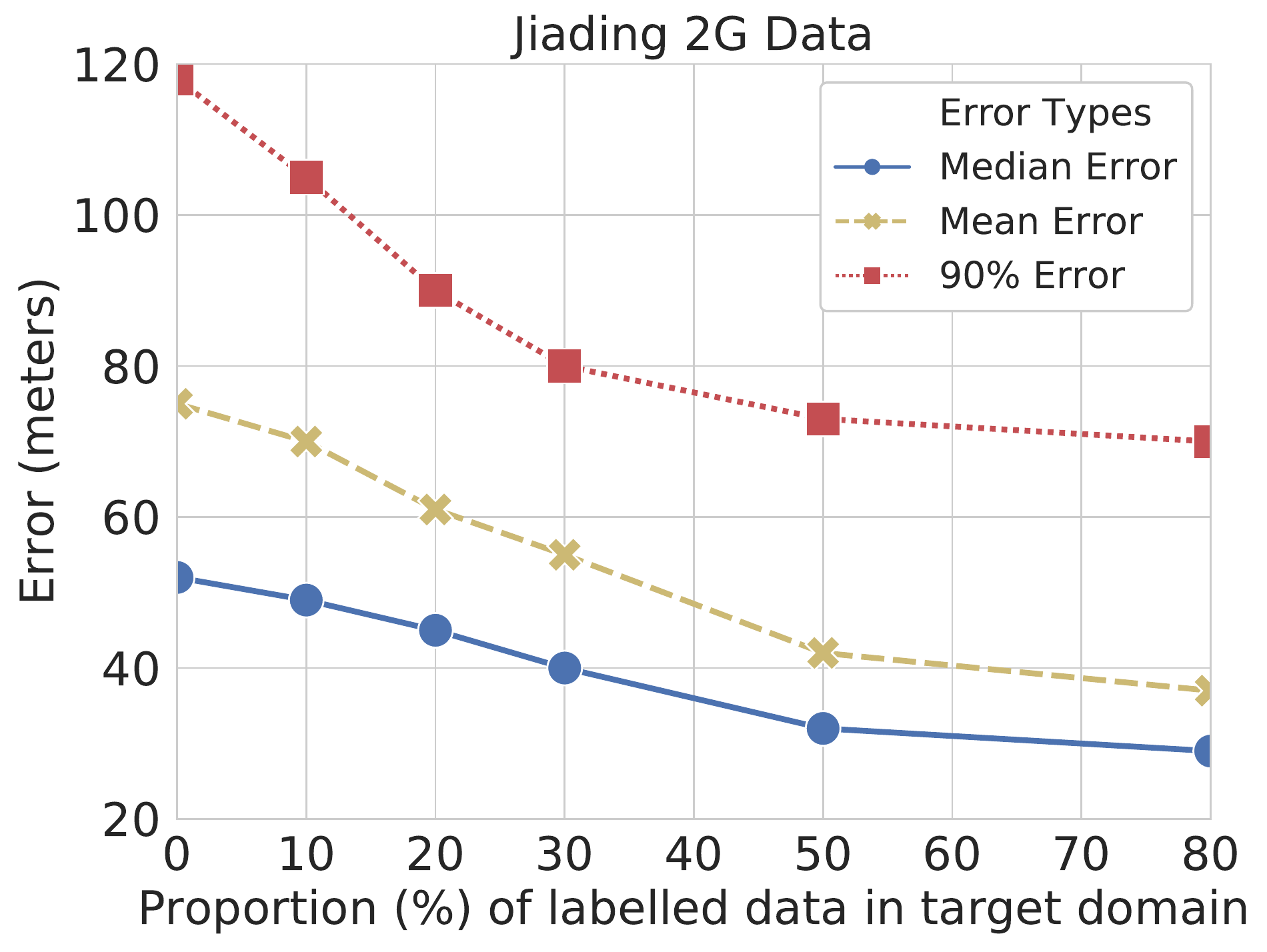}}
				\end{center}
			\end{minipage}
			&
			\begin{minipage}[t]{0.25\linewidth}
				\begin{center}
					\centerline{\includegraphics[height=1.16in]{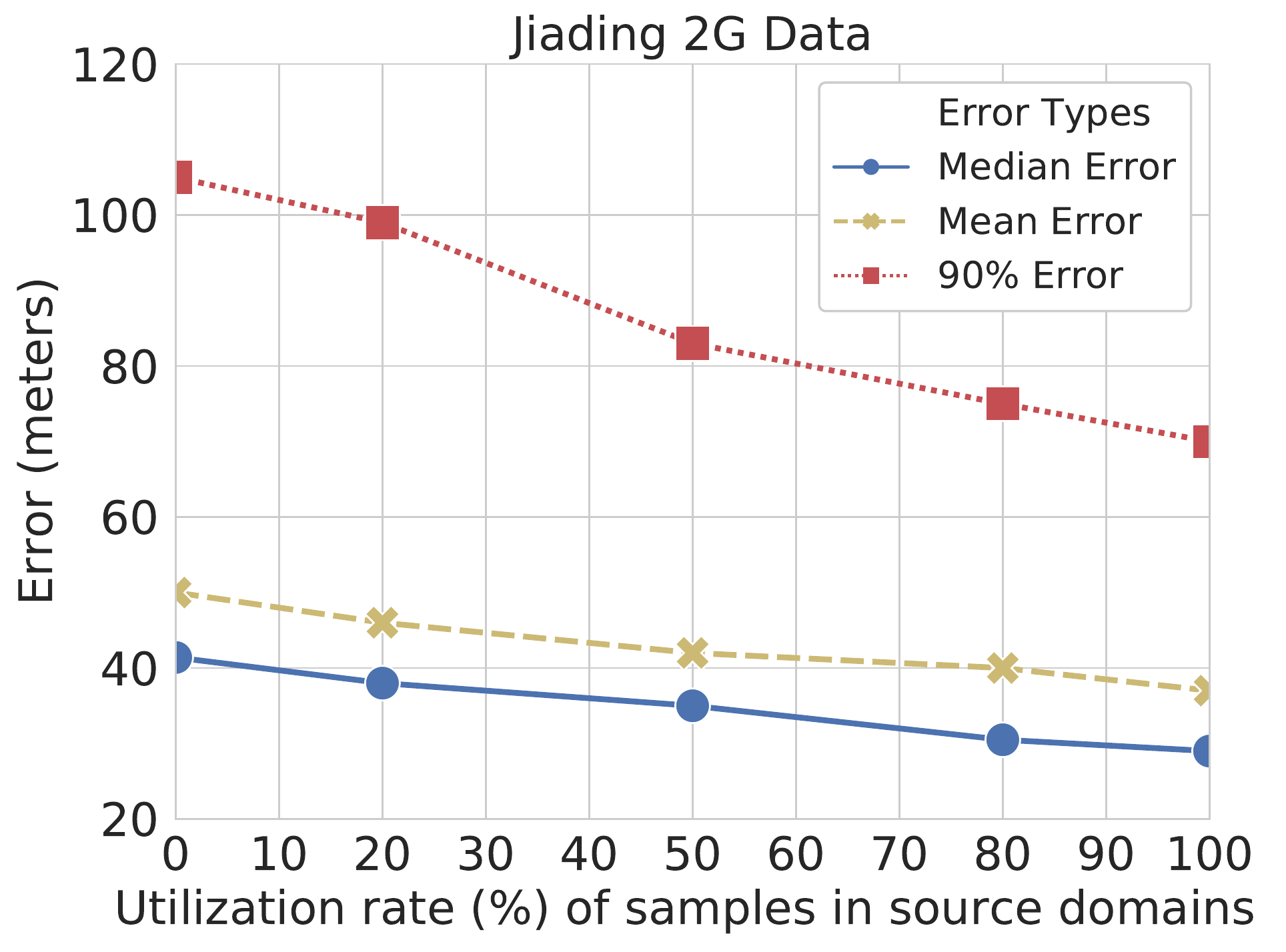}}
				\end{center}
			\end{minipage}
            &
			\begin{minipage}[t]{0.25\linewidth}
				\begin{center}
					\centerline{\includegraphics[height=1.16in]{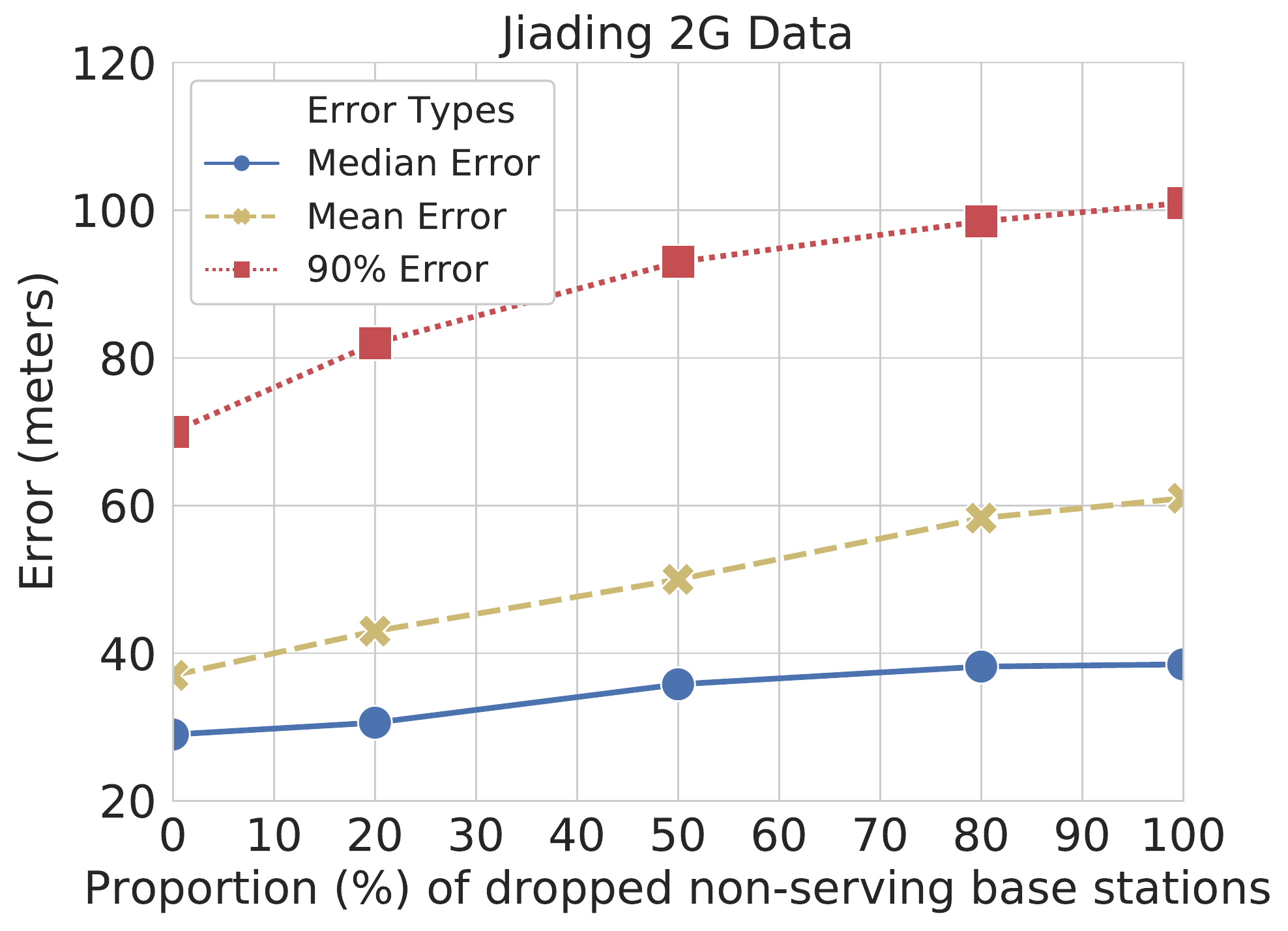}}
				\end{center}
			\end{minipage}
&
			\begin{minipage}[t]{0.25\linewidth}
				\begin{center}
					\centerline{\includegraphics[totalheight=1.16in]{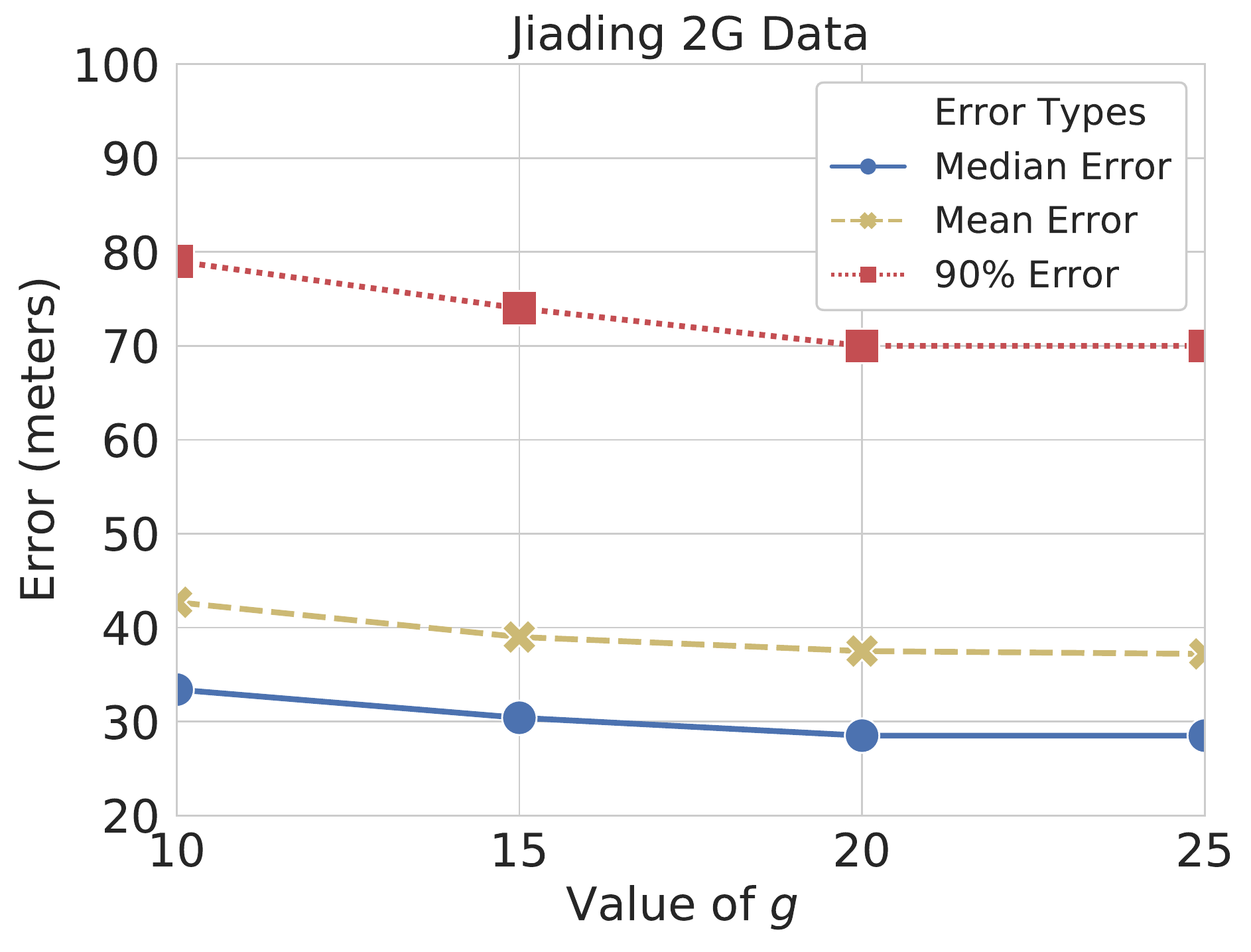}}
				\end{center}
			\end{minipage}\vspace{-3ex}
		\end{tabular}\caption{Sensitivity Study (from left to right): (a-b-c) Proportion of target samples, source samples, and dropped base stations, (d) Effect of grid size $g$.} \label{exp:sense}\vspace{-3ex}
	\end{center}
\end{figure*}

\textbf{Proportions of Target Samples:} First, by varying the proportions of data samples in target domains from $0\sim80\%$, we train \textsf{TLoc} and plot the mean, median and 90\% errors in Figure \ref{exp:sense}(a). When no data samples are used in target domains, \textsf{TLoc} has to fully leverage the trained models from source domains, leading to the highest errors. When more samples are used in target domains, the errors become gradually smaller. It is mainly because \textsf{TLoc} adapts the models originally trained on source domains towards target domains.

\textbf{Proportions of Source Samples:} Besides the samples in target domains, we also vary the proportion of source data samples from 0 to 100\% in Figure \ref{exp:sense}(b). The proportion equal to 0, i.e., the \emph{No-Transfer} approach, suffers from the highest error. More source samples lead to lower errors. Nevertheless, when comparing the sub-Figures \ref{exp:sense}(a-b), we find that \textsf{TLoc} is more sensitive to the data samples in target domains. It makes sense because \textsf{TLoc} performs the position recovery on target domains, and the data samples in target domains thus directly determine the errors of \textsf{TLoc}.

\textbf{Base Stations Density:} From Table \ref{tab:dataset}, we find that the base stations in {\emph{Jiading}} datasets (both 2G GSM and 4G LTE) are much sparser than those in \emph{Siping}. Thus, the localization errors in {\emph{Jiading}} datasets are slightly higher than those in \emph{Siping} dataset. Moreover, we note that \textsf{TLoc} builds the position recovery model by referring to serving base stations as domain centers. Thus, we randomly choose some non-serving base stations in each MR sample, and reset such base stations and associated Telco signal strength values to be empty. In this way, we drop these base stations from MR samples and vary the density of base stations in MR datasets. In Figure \ref{exp:sense}(c), the $x$-axis shows the percentage of dropped base stations and the $y$-axis gives the median error. A larger dropping rate leads to higher localization error. However, when the total number of dropped base stations rises, the prediction error does not rise sharply. This experimental result indicates that the localization precision of \textsf{TLoc} mainly depends upon serving base stations.

\textbf{Count $g$ of Divided Grids:} Recall that in Section \ref{sec:relative}, we represent the MR features $F_d()$ by grid IDs which require the division of each domain into $g*g$ smaller girds. In Figure \ref{exp:sense}(d), when the number $g$ of divided grids grows (i.e., smaller grid width/height), the error of \textsf{TLoc} first increases and then remains stable when $g>20$. The reason is as follows. A smaller $g$ (i.e., larger grid width/height) leads to more neighboring base stations within each divided grid cell, and consequently incurs the coarser-grained representation of $F_d()$. It results in higher errors. Instead, a greater $g$ divides a domain into more cells with smaller grid width/height, and thus leads to lower errors. The tuning of $g$ involves the aforementioned trade-off and we empirically set $g=20$ by default.

\subsection{Localization Visualization}\label{sec:vis}
\begin{figure}[th]
\begin{center}
\begin{tabular}{c c c}
\begin{minipage}[t]{0.25\linewidth}
\begin{center}
\centerline{\subfigure[Non-Transfer]{
    \label{fig:subfig:nt} 
    \includegraphics[width=1.1in]{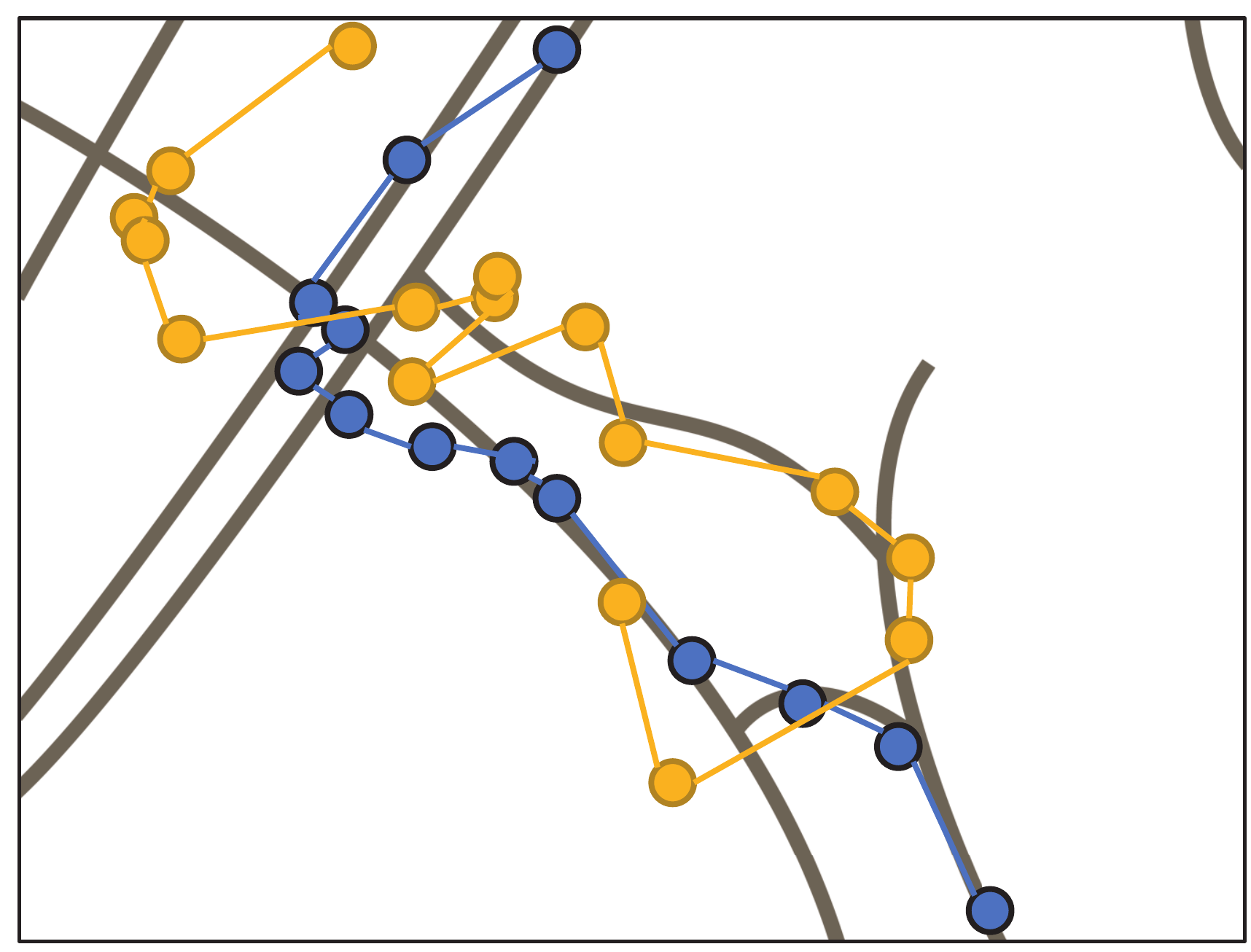}
  }}
\end{center}
\end{minipage}
&
\begin{minipage}[t]{0.25\linewidth}
\begin{center}
\centerline{\subfigure[MTL]{
    \label{fig:subfig:mtl} 
    \includegraphics[width=1.1in]{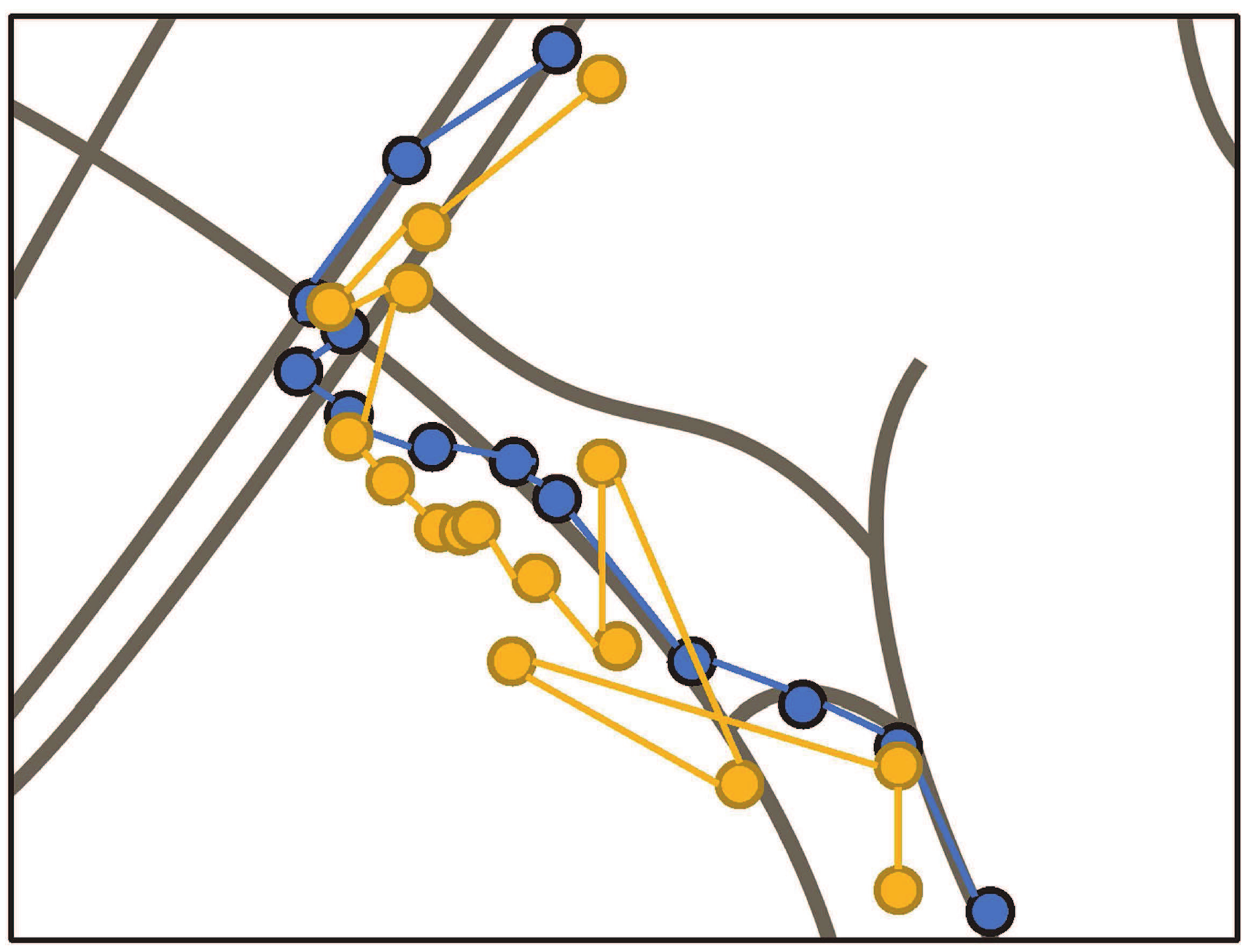}
  }}
\end{center}
\end{minipage}
&
\begin{minipage}[t]{0.25\linewidth}
\begin{center}
\centerline{\subfigure[TLoc]{
    \label{fig:subfig:ccr} 
    \includegraphics[width=1.1in]{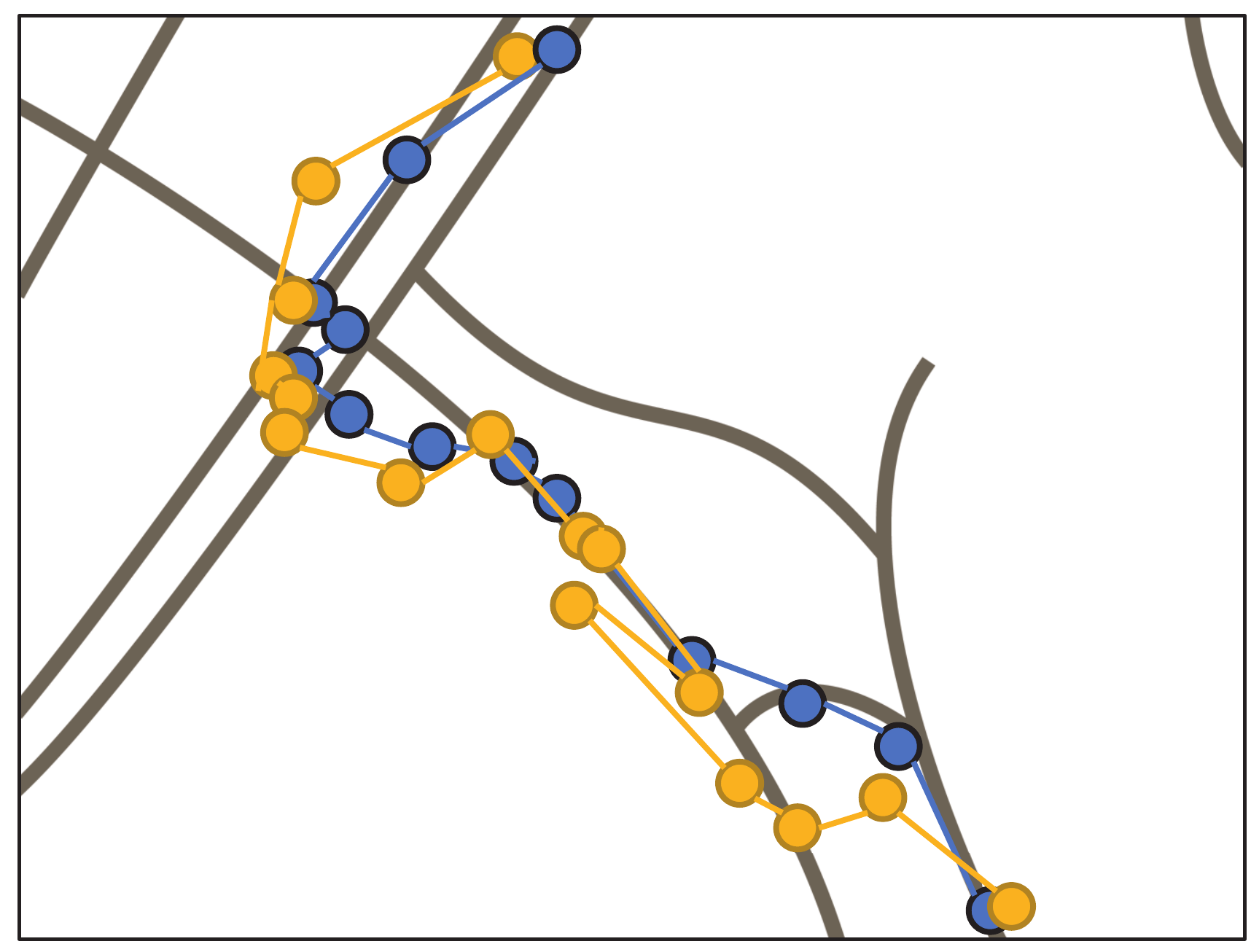}
  }}
\end{center}
\end{minipage}
\end{tabular}\vspace{-1ex}
\caption{Visualization Result. Blue: Ground truth; Orange: Predicted Position. }\vspace{-4ex}
\label{exp:vis}
\end{center}
\end{figure}

Finally we visualize the positions recovered by three RaF-based algorithms (non-Transfer, \textsf{MTL}, and \textsf{TLoc}) on a randomly selected domain in {\emph{Jiading}} 2G GSM data set. We choose these approaches is mainly because they lead to the top-3 best results. As shown in Figure \ref{exp:vis}, the blue dots represent the GPS position labels (as ground truth) and orange ones represent the prediction result of each algorithm. For each algorithm, we connect the recovered positions into a moving trajectory. By observing the two moving directions which are parallel and vertical to road segments, we find that the \emph{non-Transfer} algorithm leads to the largest significant shift in both horizontal and vertical directions. Instead, \textsf{TLoc} can achieve the least shift and the trajectory recovered by \textsf{TLoc} roughly matches the road segments.

\subsection{Discussion}\label{sec:discuss}

\textbf{Changes in Base Stations}: Recall that the relative coordination space of \textsf{TLoc} takes a serving base station as the center of a domain. Though the changes of base stations are not frequent, it is not rare that the software and/or hardware of base stations are updated. We show that how \textsf{TLoc} is adaptive to the update. First, we consider the case that a base station is moved to a new location. We then have two sets of MR samples, denoted by $S$ and $S'$, generated by the base station in the previous and new positions, respectively. To make sure that \textsf{TLoc} works, one simply way is to leverage the new MR samples $S'$ to train a new RaF regression model. Nevertheless, if the number of new MR samples $S'$ is trivial, the new model does not work very well. This is exactly the same challenge that we expect to address in this paper. To this end, we could re-use the RaF regression model which was trained by the previous MR samples $S$, and transfer this previously trained model from $S$ to $S'$. Specifically, given the model trained from $S$, we can follow the general idea of structured transfer learning (STL) in Section \ref{sec:transfer}, and re-select node thresholds in decision trees (DTs) by these new MR samples $S'$. In this way, the previous model is transferred to the new dataset $S'$. Second, due to hardware upgrade (e.g., the update from 2G base stations to 4G ones), the signal transmission power of the station might become significantly different. Given such a scenario, the new MR samples generated by the updated station do not follow the original mapping from MR features to relative positions, and \textsf{TLoc} treats the station as a completely new one and has to re-train the location model based on new MR samples $S'$.

\textbf{Upcoming 5G Network:} With the coming of 5G communication, it is highly expected that 5G base stations are much densely deployed than 2G GSM and 4G LTE stations. Nevertheless, we believe that \textsf{TLoc} can still bring benefits to Telco operators due to the following observations. \emph{1}) Nowadays a Telco operator typically maintains heterogeneous Telco networks mixed by 4G LTE, 3G WCDMA, and/or 2G GSM technologies. With the deployment of 5G network in near future, it is expected that Telco operators could still maintain heterogeneous networks. Thus, \textsf{TLoc} can still work to recover the positions of mobile devices using non-5G Telco networks. \emph{2}) Even for a 5G network, it is highly possible that 5G base stations deployed in rural areas could be much sparse than those in urban areas. \textsf{TLoc} still has chance to work well in rural areas.

\section{Conclusion and Future Work} \label{s:conclusion}
In this paper, we study the problem of Telco outdoor position recovery in the areas with insufficient geo-tagged MR samples, and design a transfer learning-based position recovery framework, namely \textsf{TLoc}. The contributions of our work include \emph{1}) the proposed relative coordinate space to represent MR features and positions, \emph{2}) the distance metric to measure the similarity of domains, and \emph{3}) a transfer learning-based position recovery framework by adapting the STL approach. Our extensive evaluation validates that \textsf{TLoc} outperforms two state-of-the-art methods (\textsf{CCR} and \textsf{NBL}) and the variants of \textsf{TLoc}. As \textsf{TLoc} is a first stepping stone to explore transfer-learning for Telco outdoor position recovery, the promising results motivate the following future work, e.g., deep neural network (DNN)-based position recovery and transfer learning in DNN \cite{YosinskiCBL14}.

\bibliographystyle{abbrv}
\bibliography{tloc}

\vspace{-4ex}
\end{document}